\documentclass{article}
\usepackage{PRIMEarxiv}
\usepackage[utf8]{inputenc} 
\usepackage[T1]{fontenc}    
\usepackage{hyperref}       
\usepackage{url}            
\usepackage{booktabs}       
\usepackage{amsfonts}       
\usepackage{nicefrac}       
\usepackage{microtype}      
\usepackage{lipsum}
\usepackage{fancyhdr}       
\usepackage{graphicx}       
\usepackage{float}
\usepackage{subfiles}
\usepackage{amsmath}
\usepackage{bm}
\usepackage{cite}
\usepackage{units}
\usepackage{wrapfig}
\usepackage{indentfirst}
\usepackage{multirow}
\graphicspath{{media/}}     

\title{Design and Development of a Locomotion Interface for Virtual Reality Lower-Body Haptic Interaction}

\author{
  An-Chi He*, Jungsoo Park, Benjamin Beiter, Bhaben Kalita and Alexander Leonessa\\
  Terrestrial Robotics Engineering and Controls (TREC) Laboratory\\
  Virginia Tech, Blacksburg, VA 24060, USA \\
 \\
  \textit{*Corresponding author:} \texttt{anchihe@vt.edu} \\
}
\def\v{_{\rm v}}
\def\d{_{\rm d}}

\begin{document}
\maketitle
\setlength{\parindent}{20pt}

\begin{abstract}
This work presents the design, build, control, and preliminary user data of a locomotion interface called ForceBot. It delivers lower-body haptic interaction in virtual reality (VR), enabling users to walk in VR while interacting with various simulated terrains. It utilizes two planar gantries to give each foot two degrees of freedom and passive heel-lifting motion. The design used motion capture data with dynamic simulation for ergonomic human-robot workspace and hardware selection. Its system framework uses open-source robotic software and pairs with a custom-built power delivery system that offers EtherCAT communication with a 1,000 Hz soft real-time computation rate. This system features an admittance controller to regulate physical human-robot interaction (pHRI) alongside a walking algorithm to generate walking motion and simulate virtual terrains. The system's performance is explored through three measurements that evaluate the relationship between user input force and output pHRI motion. Overall, this platform presents a unique approach by utilizing planar gantries to realize VR terrain interaction with an extensive workspace, reasonably compact footprint, and preliminary user data.
\end{abstract}


\section{Introduction}
\par Virtual reality technology has received notable advancements in the past decades~\cite{zhang2020,ali2017,luis2020}, and there are increasingly more fields that utilize this technology for various purposes, offering significant benefits across education~\cite{martin2018,app11062879}, healthcare~\cite{sik2017virtual}, entertainment~\cite{Vorderer2021}, engineering~\cite{wang2018critical, berni2020applications}, and others. VR offers a simulated world through various senses, and the most common forms of VR sensory feedback are visual and auditory, which have proven effective at creating immersive experiences. In recent VR applications, the addition of haptic feedback has been shown to improve this immersion further, providing measurable improvements for simulating realistic scenarios such as skill training~\cite{Collaco2021, Gani2022}, education~\cite{Chrysanthakopoulou2021, Edwards2019}, and rehabilitation~\cite{Bortone2018}. However, most haptic devices focus on providing feedback to the upper body; it is rarely incorporated with lower-body interaction. Integrating haptic feedback and lower-body interaction can potentially improve the comfort and effectiveness of navigation in VR. The most common virtual environment (VE) navigation methods often use controllers, joysticks, or sliding platforms. While effective, these devices usually lack bodily motion or cannot simulate more than a simple flat ground. The ability to display computer-generated terrains with haptic interaction can broaden VR applications to simulate more extensive scenarios, allowing users to traverse virtual terrain with natural body movements. This paper presents a VR haptic device focusing on such lower-body interaction, providing walking motion while allowing users to interact with simulated terrains.

Even with VR technology being used across various fields, VR-induced discomfort has become prevalent as a notable concern documented in several studies~\cite{chattha2020motion,kolasinski1995simulator,lo2001cybersickness}, hindering long-term uses and limiting broader utilization of VR. The sensory mismatch between visual and bodily senses is one of the key factors that cause VR motion sickness~\cite{matt2021,NG2020101922}. This mismatch comes from how users navigate in the VE: the predominant navigation method often involves hand-held controllers, often with a push-to-move joystick or a point-to-teleport mechanism, which neglects bodily motion. Human brains rely on a combination of sensory inputs to process information and anticipate the next movement. These sensory inputs include visual, vestibular, haptic, and auditory cues. Continuous visual stimulation without accompanying bodily sensory inputs can cause cognitive stress, thus fatiguing our brains~\cite{Heo2020}. Enabling bodily movement to reduce sensory conflict is one of the potential solutions to reduce VR motion sickness~\cite{Chang2020}. Research has shown that incorporating walking motion for VR navigation is preferred over joy-stick or point-to-teleport mechanism; it increases the user's VE spatial knowledge, enhances the sense of presence, and can reduce VR-induced motion sickness~\cite{Langbehn2018}. This result agrees with other studies~\cite{Choi2020,Mayor2021,chattha2020motion}, suggesting that a VR locomotion technique using bodily motion is desirable for improving the VR experience. The exploration of such locomotion techniques in VR has been gaining attention in recent years~\cite{Cherni2021,Kitson2017,Bovim2020}. These locomotion techniques often involve different types of locomotion interfaces, devices, or systems that enable users to navigate virtual environments. It serves as the bridge between the user's physical actions and their digital world movements, allowing users to explore and navigate a simulated environment and to be fully immersed in the virtual world. However, standard VR locomotion devices often cannot display the variable terrains our human world contains, such as slopes, stairs, and bumps. Being able to represent these three-dimensional features may enable VR applications in more scenarios, such as realistic virtual training~\cite{Vukelic2023,Conges2020}.

Several existing VR locomotion techniques incorporate bodily movements for realistic experiences and reduce the risk of VR-induced discomfort. These techniques are categorized into real-walking, redirected walking, walk-in-place, and using locomotion interfaces~\cite{Boletsis2017,Bozgeyikli2019,Cherni2020}. The real-walking method tracks the user’s head-mount-display position within a limited physical space, which most commercial VR devices such as HTC VIVE and META Oculus use. This method allows natural bodily movement but has the limitation of a constrained physical space to move within. The redirected walking method is then built upon the real-walking method to resolve its limitation. Redirected walking~\cite{Razzaque2001,Matsumoto2016} subtly manipulates the user’s sense of direction and movement in the virtual environment, allowing them to explore large virtual spaces while walking in a smaller physical area. It is worth noting that there is a threshold for how much manipulation can be done before users can notice~\cite{Steinicke2010}. Walk-in-Place~\cite{Slater1995} is a technique where users emulate walking movements, such as lifting their legs while remaining in the same position. The VR system tracks these movements and translates them into forward, or turning motion within the virtual environment, giving the user the sensation of walking without the need for extensive physical space or the risk of collisions. Locomotion interfaces~\cite{Iwata2000} refer to devices or systems that are built to enable users to navigate and interact with VR using bodily motion. A treadmill is a prime example of such a device. These devices aim to allow natural bodily movement while maintaining users in the same physical location. Locomotion interfaces are often built with a specific aim, each with its own strengths and limitations. Many locomotion interface designs can provide haptic interaction or simulate virtual terrains that other previously mentioned techniques are incapable of, providing a much more immersive experience. However, the apparent downside is the development cost to build a robotic system while allowing natural bodily movements.

Currently, the three major types of locomotion interface designs are (i) sliding-platform, (ii) treadmill-based locomotion interface, and (iii) robotic foot-platform locomotion interface. The sliding-platform designs, such as the KAT Walk C2, Virtuix OMNI, or Virtualizer ELITE~\cite{Hager}, are the most common commercially due to their simple design and low cost. These VR locomotion devices incorporate a low-friction surface to facilitate smooth walking movement, allowing users to glide their feet over the ground surface to mimic walking or running motions while holding them in a constant physical location with a torso harness. Treadmill-based locomotion interfaces are also meant to keep a user in a constant physical location, but with a variable speed treadmill belt the user walks on to emulate walking in VR. It allows an uninterrupted walking experience that is more similar to a natural walking gait than a sliding platform, but has a higher cost and complexity due to the active motors. There are advanced treadmill designs that can display variable ground stiffness~\cite{barkan201400, Skidmore2015, hernandez201800} or add an extra degree of freedom (DOF) to be a 2DOF planar treadmill~\cite{Pyo2018, Hollerbach2000}. However, a prominent limitation is the inability of both types of designs to display three-dimensional ground geometries, leaving a significant gap in achieving realistic VR locomotion.

The third type of locomotion interface is robotic foot platforms. This design incorporates robotic systems where the user stands with each foot on an end-effector that can move independently to display any simulated terrain through force or motion feedback at the foot. Several such devices have been created. Schmidt et al.~\cite{schmidt2005hapticwalker} introduced the haptic walker for rehabilitation applications. While it worked to allow a continuous walking experience, its substantial size restricts it from being used in space-limited research labs or rehabilitation clinics. Boian et al. proposed a design that employs two Stewart platforms~\cite{boian2005}, offering a comprehensive 6 DoFs on each foot but with the limitation of a restricted workspace in X-Y plane (transverse plane) motion. The Gait Master~\cite{Iwata20010} comprises two 2 DoF parallel drive platforms as a moving ground under the user's feet. In contrast to other designs, the user's feet are not directly attached to the foot platforms instead the user's shoe position is tracked and the platforms are controlled to always be underneath the foot. Some concerns of this type of design comes from position sensing accuracy, and potential injury foot platform misplacement. Yoon et al. extended the shoe position tracking mechanism and incorporated a turning feature, allowing users to navigate in VEs freely~\cite{yoon2010planar}. Another design by Yoon et al.~\cite{yoon2010} synchronizes upper and lower limbs for rehabilitation. The most recent design is carried out by the BiONICS Lab at UCLA (University of California, Los Angeles), which utilizes four robotic arms to deliver haptic interactions for the upper and lower limbs. So far, there has been no publication on this device yet. The robotic foot platform design has an inherent programmable nature suitable for incorporating haptic feedback and displaying simulated terrains, making it an ideal option for locomotion devices. However, no prior device has been able to address all of the challenges presented by a robotic foot platform at once, such as having a limited workspace, substantial device footprint, body turning motion, comprehensive 6-DOF on the feet, or smooth physical Human-Robot Interaction (pHRI). Each prior device is purposefully built and has its pros and cons. 

The robotic foot platform design type aligns well with our goal of creating a platform for lower-body interaction and VR terrain simulation. No previously designed devices for lower-body haptic interaction are commercially available. Additionally, previous studies~\cite{yoon2010,yoon2010planar,Iwata20010,schmidt2005hapticwalker,boian2005} have not provided enough data to guarantee high performances while testing more complex haptic control algorithms, leaving a significant research gap. These studies either do not include preliminary user data or have users walking slowly in a small workspace. This work builds upon these previous designs by contributing a detailed development process for a locomotion interface with a larger workspace and velocity tracking capability, along with experimental measurements to verify device performance. The purpose of the device is to be a study platform for lower-body haptic interaction. This paper shares a system framework that can be applied to different research groups, primarily using off-the-shelf components and open-source robotic software. In this work, we present a novel robotic foot-platform-based locomotion interface, introduce its design, build, and control, and demonstrate its results. It uses motion capture data to design a system workspace sufficient for walking motion without a large device footprint. This system features high-power industrial motors to support the user's weight and creates responsive pHRI. The system contains two planar gantries, each with a custom-designed foot platform that provides two active DoFs and a passive heel-off motion for each foot. A walking algorithm is presented which generates walking motion based on different gait phases. The algorithm can simulate three types of terrain to demonstrate its capability, which are 1. flat ground, 2. stairs, and 3. arbitrarily uneven terrain. It features an admittance controller to regulate human-robot interaction, measuring the interaction force to provide robotic motion. We propose a system framework for a 1,000 Hz soft real-time computation and EtherCAT communication. 

The rest of the paper will be laid out as follows: Section 2 provides an overview of the design and hardware. Section 3 presents its communication framework, addressing every major component of this system. Section 4 presents an admittance controller that regulates pHRI, and Section 5 shows the walking algorithm that provides walking interaction and simulates different types of terrains. Section 6 documents the system's performance through three measurements, and lastly, Section 7 contains the conclusion and future direction for this study.  

\section{System Design}\label{sec2}
\begin{figure}[b]
\centering
\includegraphics[width=0.9\textwidth]{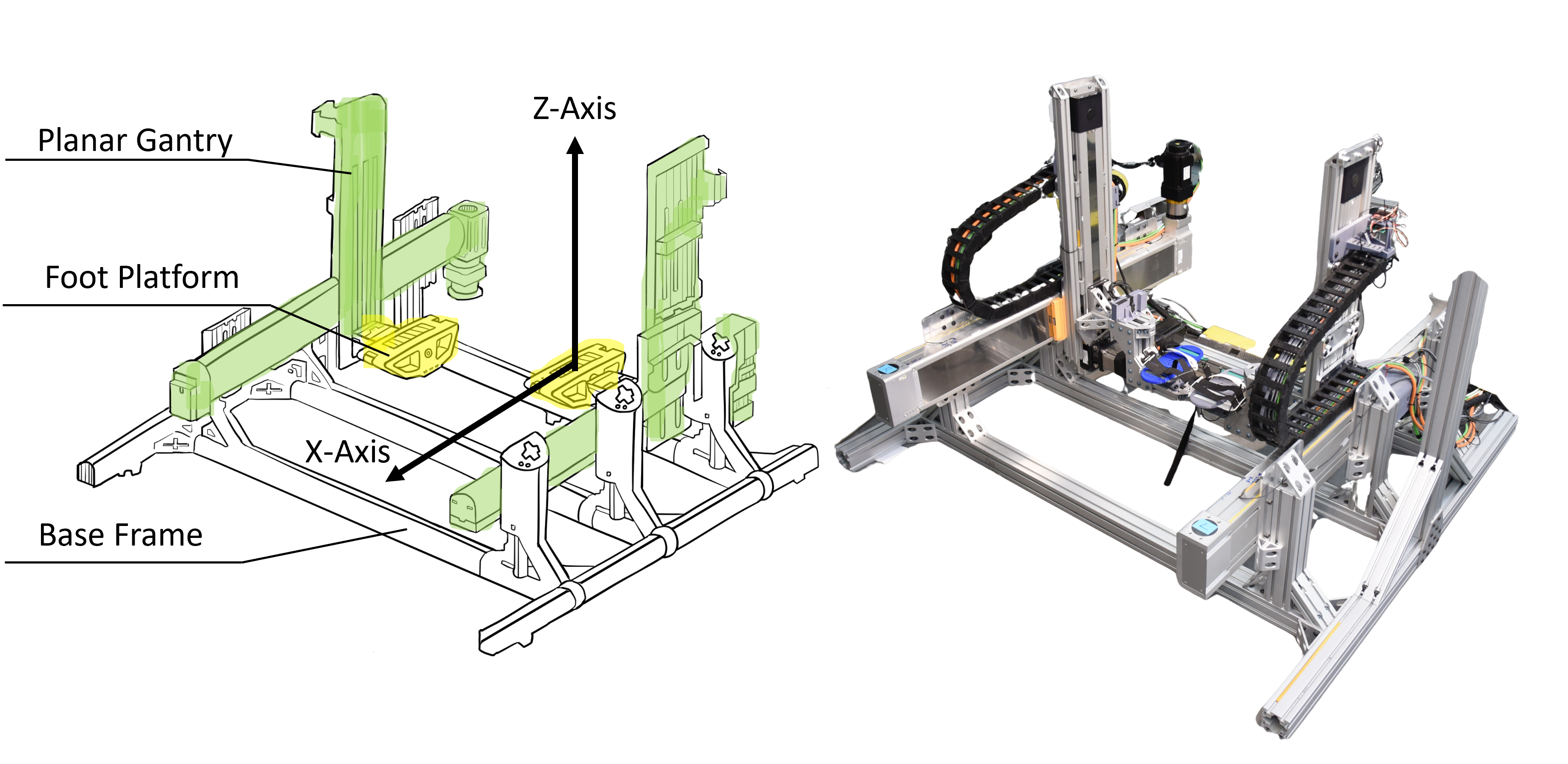}
\caption{A concept sketch of the presented locomotion interface alongside the actual device. The locomotion interface features two linear gantries, a pair of foot platforms, and a base frame.}
\vspace{-10pt}
\label{fig:overallStructure}
\end{figure}
The presented locomotion interface, ForceBot, is designed to enable users to traverse VR terrains using a robotic system. This system contains two planar gantries as shown in figure~\ref{fig:overallStructure} marked by green color. Each planar gantry has a foot platform connected to it as a physical user interface for users to attach their feet to the system, marked by yellow in figure~\ref{fig:overallStructure}. Each foot platform can move independently in the sagittal plane, labeled the X and Y-axis. These foot platforms also allow passive heel-lifting motion to improve the walking interaction despite having a limitation in toe-lifting (dorsi-flexion) motion. This system provides a treadmill-like walking experience by sliding the platform along a virtual ground during the stance phase; changing the constraints of the robotic motion allows for simulating different virtual terrain. During the swing phase, the foot platform simply follows the motion of the user's foot. An admittance controller regulates this physical human-robot interaction, translating the interaction force into robotic motion. 

Currently, there is no robotic foot platform locomotion device that is commercially available. However, the need for a high-performing, reliable platform for developing lower-body haptic interaction controllers motivates this design. The presented device balances workspace and footprint size, control performance, and cost-effectiveness. The proposed design offers a large robotic workspace compared to a Stewart platform-based design~\cite{boian2005}, it also has a smaller footprint size than some previous devices~\cite{schmidt2005hapticwalker,yoon2010planar}. This device has utilized off-shelf components to reduce development costs, and speed up development time. The limitations of this more compact, cost-effective device are uni-directional walking and unactuated ankle motion, which prevents users from navigating in VR freely. Despite its limitations, the device is still competent in being a platform for terrain simulation study. The rest of this section documents the design process, comprehensively covering general design requirements, system design, hardware selection, foot platform design, and an electric power delivery system. 

\vspace{0pt}
\begin{figure}[t]
\centering
\includegraphics[width=0.4\textwidth]{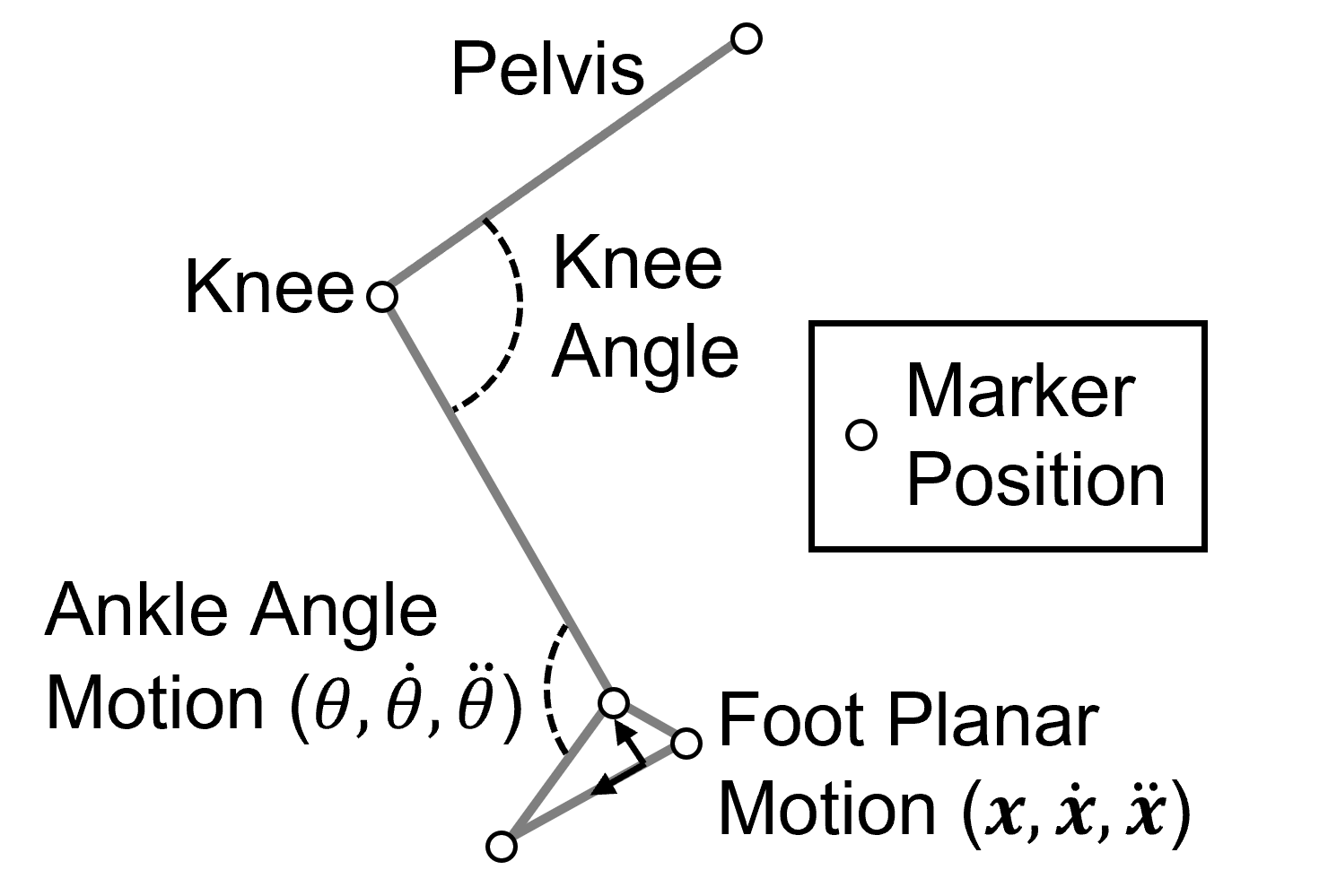}
\caption{Motion capture marker positions to obtain walking motion data for dynamic simulation.}
\vspace{-10pt}
\label{fig:moCapDia}
\end{figure}
 
\subsection{General Design Requirements from Human Motion Capture Data}
Since locomotion interfaces often require users to physically interact with the robotic system, which may raise safety concerns and make it challenging to recruit external subjects for early-stage development. We present a design process featuring an internal member for maintaining consistency across prototyping and pilot testing. The same design process can be adapted by other developers to tailor the device to their needs. The first step of the design process is defining the general requirements to achieve walking motion on a robotic platform. These requirements include actuator force/torque, actuator velocity, acceleration, and workspace size. Motion capture data of a subject weighing 47 kg walking at a consistent speed of 1.2 m/sec (the average human walking speed~\cite{laplante2004continuing}) is used to align the human-robot workspace and equipment requirements. Figure~\ref{fig:moCapDia} illustrates the motion capture marker positions on the subject's leg. The position of each point is gathered during walking, allowing for the reconstruction of full leg motion during the entire dataset. The motion capture data shows a maximum step length and foot clearance of 0.67 meters and 0.14 meters, respectively, along with estimated foot velocity and acceleration as shown in table~\ref{tab:table1}. 

\begin{figure}[!t]
\centering
\includegraphics[width=.9\textwidth]{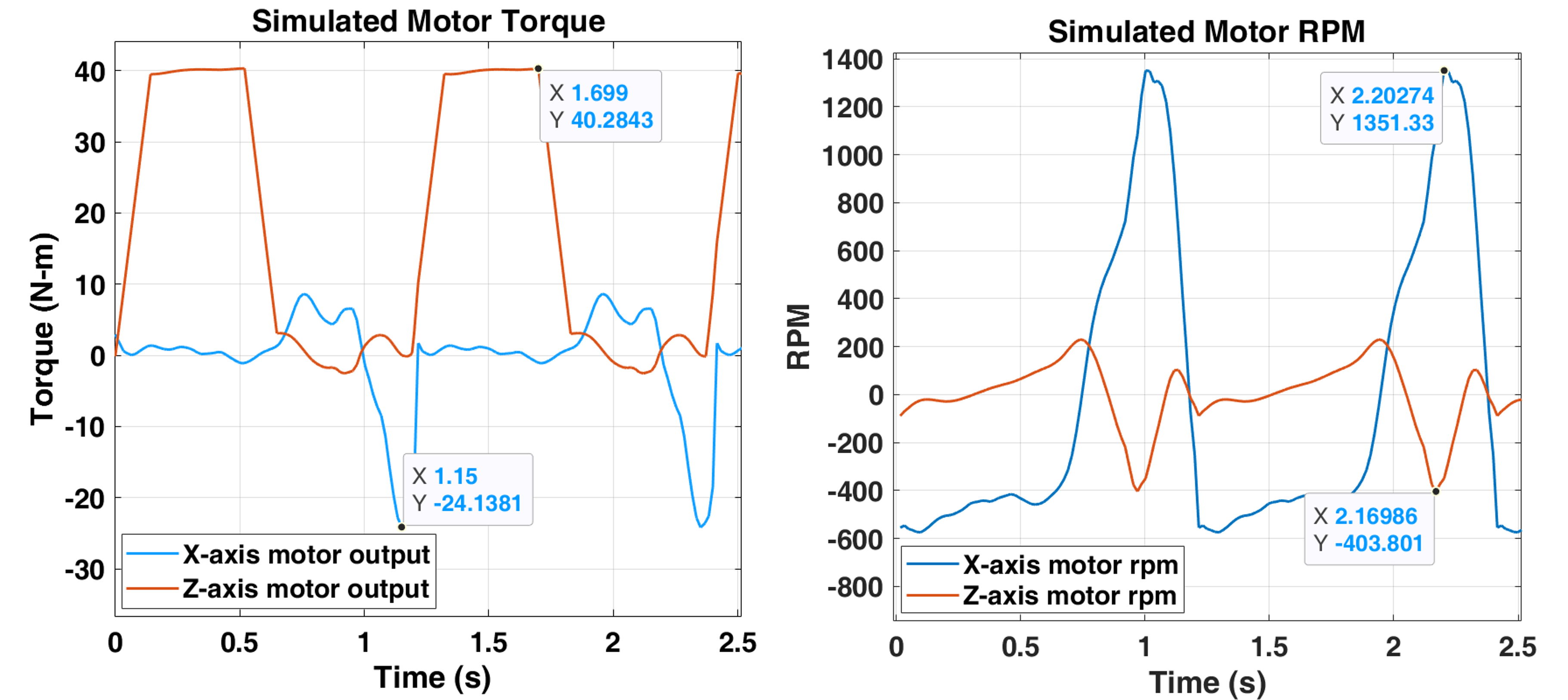}
\caption{The simulated motor torque and rotation speed that are based on motion capture data of walking. Right: Simulated motor torque for walking motion. Left: Simulated motor RPM for walking motion.}
\label{fig:motorTorque}
\end{figure}

The device workspace is designed to be larger than the footstep
length and clearance observed in the motion capture data. The chosen workspace has a 1-meter horizontal (X-axis) and a 0.5-meter vertical (Z-axis) range of motion. Once the workspace is determined, the motion capture data is used in a dynamic simulation to obtain the required motor torque and velocity to finalize the hardware selection. The simulation is done in MATLAB Simulink with a Simscape Multibody toolbox. The simulation model includes multiple actuator candidates and their hardware specifications for design evaluation. Specifications such as actuator inertia, motor torque to actuator force conversion, and structural load capability are considered in the simulation. Due to the lack of direct ground reaction force measurement in our motion capture data, an approximated ground reaction force is applied artificially in the simulation by adding an external force load to simulate user weight, with a maximum applied load to simulate a 90 Kg user. The simulated weight is applied on the foot platform during the stance phase and removed during the swing phase. A ramp function is added to smooth the weight transition between the stance and swing phase. Figure~\ref{fig:motorTorque} shows the simulated motor torque and velocity that would be required to support the user during walking motion on the gantry-style platform. The simulated maximum motor torques $|\tau|$ are approximately 24.1 N-m and 40.3 N-m for the X and Z-axis, respectively. The simulated maximum motor speeds $|\omega|$ are around 1351 RPM for the X-axis and 403 RPM for the Z-axis. The X-axis requires lower torque but higher velocity, while the Z-axis requirement is the opposite. These characteristics are due to the X-axis having to cover a large forward movement during the swing phase, while the Z-axis covers a shorter movement for foot clearance but requires a higher motor torque to support the user's weight during the stance phase. These general requirements are listed in table~\ref{tab:table1} alongside with the selected motor specifications for comparison. The next section further explains the equipment selections.

\begin{table}[h]
\centering
\caption{The motion capture data, simulation result, and selected equipment. Both the X and Z axes use the same motor, each equipped with a gear reduction ratio to adjust for different requirements.}
\begin{tabular}{ l c c }
\hline
\textbf{Motion Capture Data} & \multicolumn{2}{c}{\textbf{Maximum Value}}\\
\hline
Foot Step Length $x$ & \multicolumn{2}{c}{0.67 m}\\ 
Foot Clearance $z$ & \multicolumn{2}{c}{0.14 m}\\ 
Foot Velocity ($|v_{x,\rm max}|$, $|v_{z,\rm max}|$) & \multicolumn{2}{c}{$(2.92, 0.86)~\nicefrac{\text{m}}{\text{s}}$}\\
Foot Acceleration ($|a_{x,\rm max}|$,$|a_{z,\rm max}|$) & \multicolumn{2}{c}{$(26.75, 15.06)~\nicefrac{\text{m}}{\text{s}^2}$}\\
\hline
\textbf{Simulation Result} & \textbf{X-axis} & \textbf{Z-axis}\\ 
\hline
Simulated Max Motor Torque $|\tau|$ & 24.1 Nm & 40.3 Nm\\ 
Simulated Max Actuator Forces & 1,179 N & 1,058 N\\
Simulated Max Motor Speed $|\omega|$ & 1,351 RPM & 403 RPM\\
\hline
\textbf{Selected Motor: Omron R88M-1L2K030TS2} & \textbf{X-axis} & \textbf{Z-axis}\\
\hline
Gear Reduction Ratio & 3 & 10\\
Rated Motor Torque (After Gear)& 19.11 Nm & 63.7 Nm\\
Rated Actuator Force (After Gear)& 932.2 N & 1676.3 N\\
Momentary Maximum Torque (After Gear)& 57.3 Nm & 191 Nm\\
Momentary Maximum Force (After Gear)& 2,795 N & 5,078.9 N\\
Maximum Rotation Speed (After Gear)& 1,667 RPM & 500 RPM\\

\hline
\end{tabular}
\label{tab:table1}
\end{table}

\begin{figure}[t!]
\centering
\includegraphics[width=0.65\textwidth]{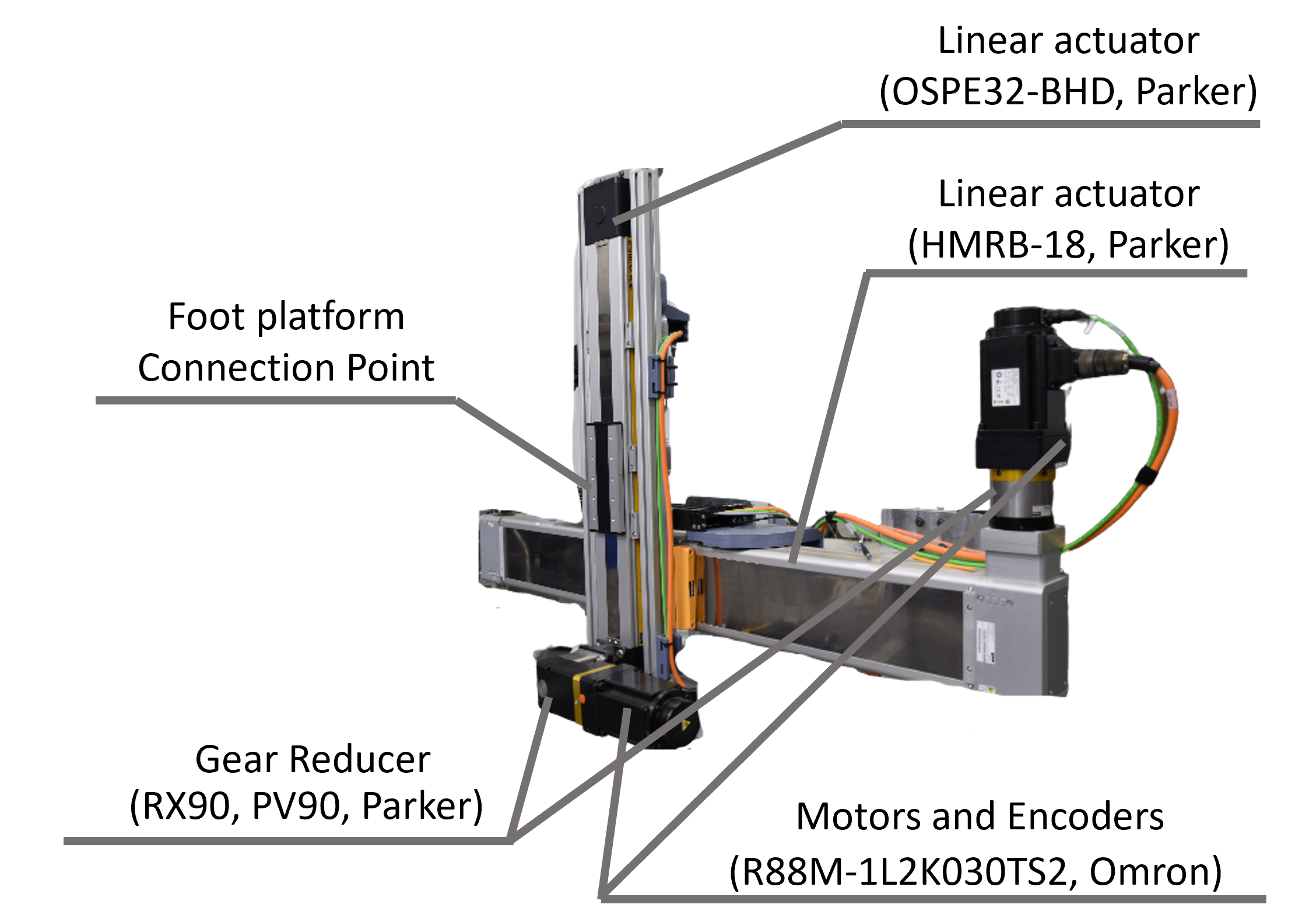}
\caption{A detailed composition of the planar gantry system. The gantry actuates X and Z-axis motion to provide 2 DoF for each foot, featuring linear rails (Model HMRB-18, OSPE32-BHD by Parker), motors (Model R88M-1L2K030TS2 by Omron), gear reducers (Model RX90, PV90 by Omron), and a connection point for the foot platform.}
\label{fig:planarGantry}
\end{figure}

\subsection{Gantry System Design}
Once the general design requirements are obtained from dynamic simulation, the equipment of the gantry system is selected accordingly. Each gantry contains two linear actuators (Model HMRB-18, OSPE32-BHD by Parker) that provide two DoFs for the user's feet, as shown in figure~\ref{fig:planarGantry}. The two gantries are connected by a base frame constructed from 80 mm square T-slotted aluminum extrusion, allowing for easy adjustment in spacing between them. Each gantry carries a foot platform, which is where the user attaches their feet to interface with the system. Limit switches are installed at each end of the actuator to prevent collisions. Each linear actuator is driven by a 2,000-watt servo motor (Model R88M-1L2K030TS2 by Omron) paired with a gear reducer (Model PV90 on the X-axis, RX90 on the Z-axis, by Parker). These motors provide 6.37 Nm rated torque and 19.1 Nm momentary maximum torque with a torque constant of 0.56 $\nicefrac{\textrm{Nm}}{\textrm{A}}$, a mechanical time constant of 0.5 $ms$ and a nominal rotation speed of 3,000 $\nicefrac{\textrm{rev}}{\textrm{min}}$. All motors in the system are identical to simplify the development, using the same motor driver, communication protocol, and encoder. The X-axis motor is paired with a 3:1 gear reduction ratio, and the Z-axis motor uses a 10:1 gear reduction ratio (based on the simulation results in Table~\ref{tab:table1}). Each motor has a high-resolution 23-bit absolute encoder and is controlled by an Omron motor driver (Model R88D-1SN20H-ECT by Omron) that provides motor velocity control. These motor drivers receive commands from a central controller computer through the EtherCAT communication protocol. 

\subsection{Foot Platform Design}
\begin{figure}[b]
\centering
\includegraphics[width=0.71\textwidth,trim={0cm 1.5cm 0 1.2cm},clip]{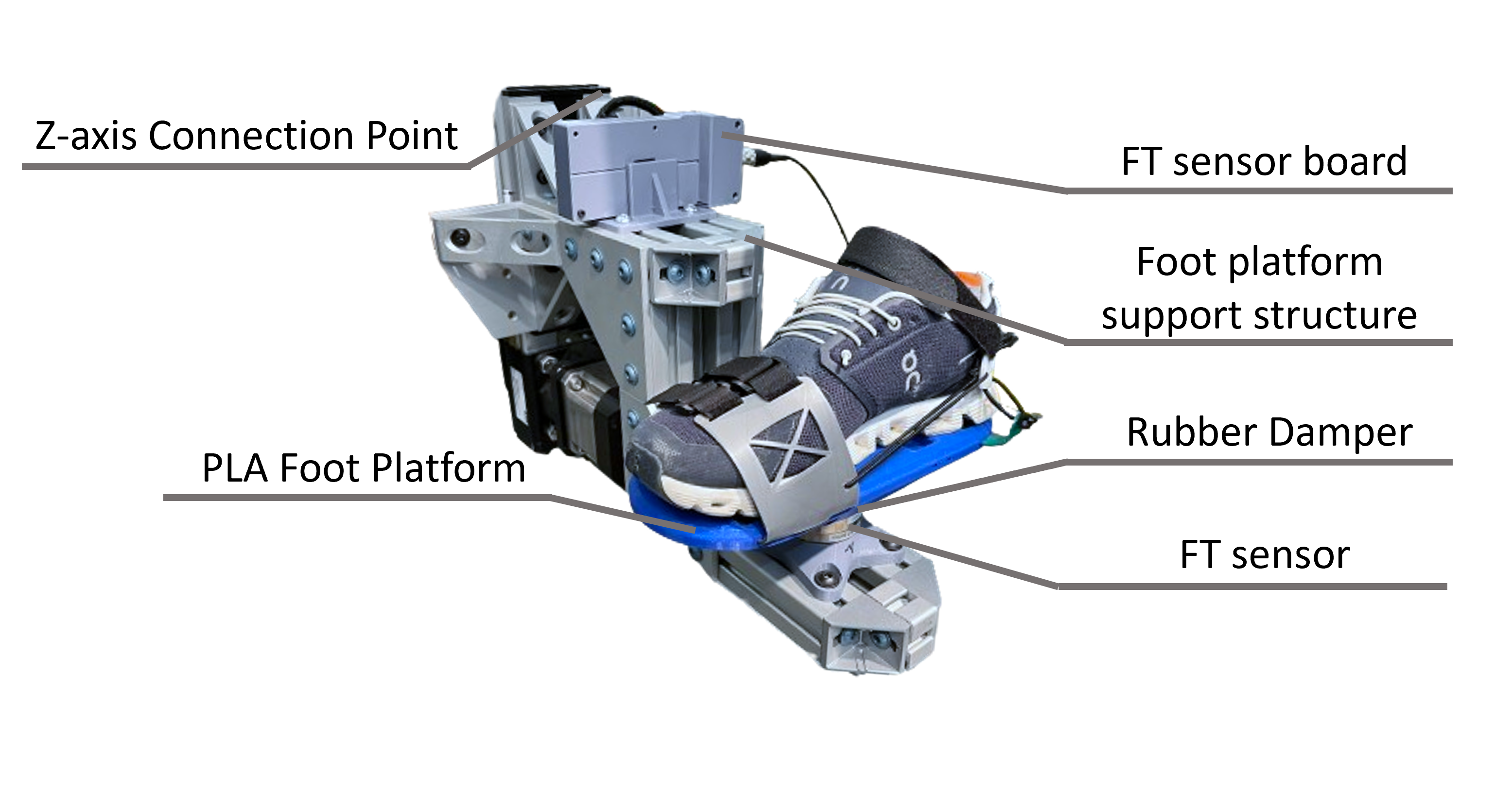}
\vspace{-10pt}
\caption{The overall foot platform design contains a support frame made of T-slotted aluminum extrusion, a PLA foot platform base, TPU foot straps, a force/torque sensor (Model MINI-58 by ATI), and a force sensor board. A user's foot is attached to the platform through TPU straps and hook-and-loop fasteners. A rubber dampening layer is implemented in between the foot platform and force sensor to reduce the negative effect of a stiff force sensor.}
\label{fig:footplatformstruc}
\end{figure}

\begin{figure}[t!]
\centering
\includegraphics[width=0.75\textwidth,trim={0cm 1.45cm 0 0.2cm},clip]{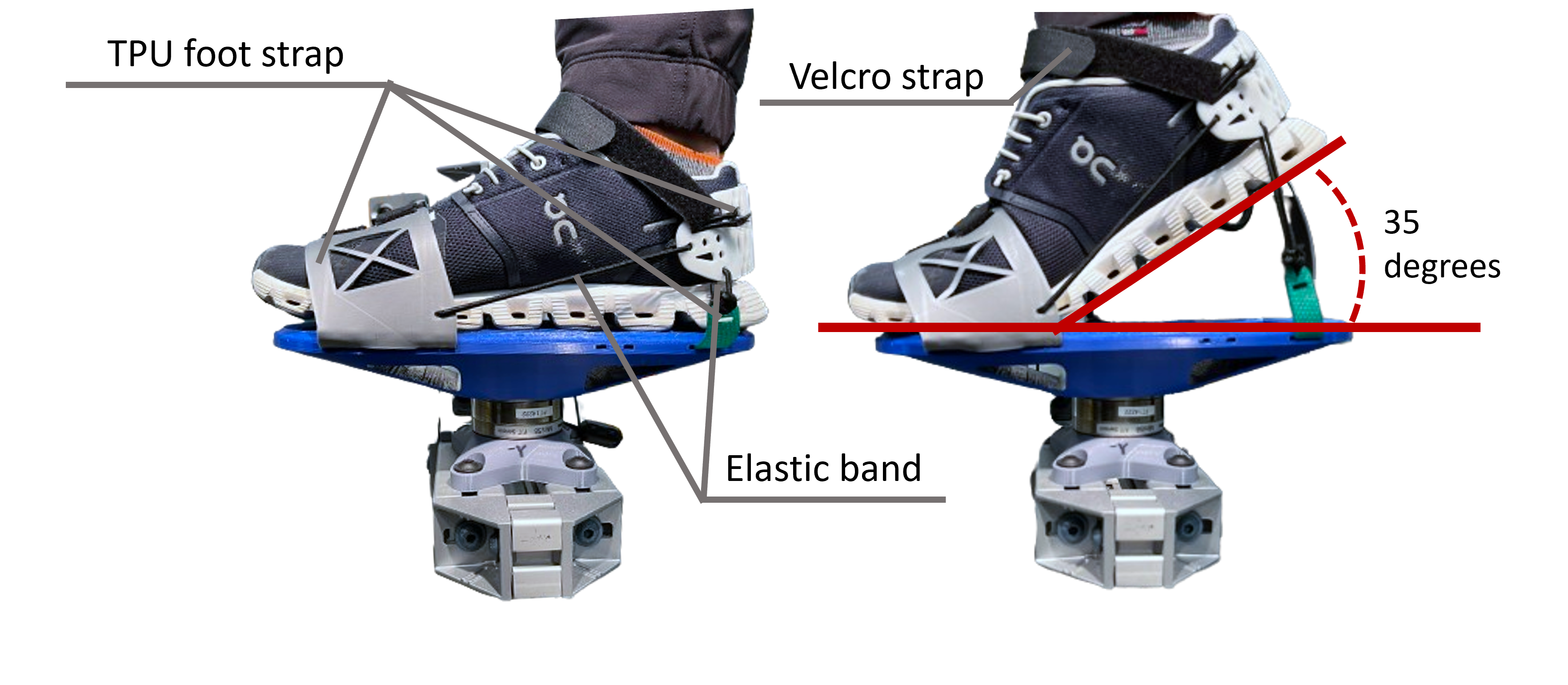} 
\caption{The attachment mechanism and passive heel-lifting of the foot platform that uses 3D-printed TPU bands with
elastic stings. Left: An image of a user standing on the foot platform. Right: An image
showing the passive heel lifting motion up to 35 degrees.}
\vspace{-10pt}
\label{fig:footmounting}
\end{figure}

The goal of the foot platform design, shown in figure~\ref{fig:footplatformstruc}, is to provide a physical interface between the user and the locomotion device that doesn't slip during any motion, allows for heel-lifting, and can sense all force interactions between the device and the user. The foot platform base, the part the foot attaches to, is 3D-printed, made of polylactic acid (PLA) plastic. This material provides the required structural strength while being lightweight and simple to print, allowing for future design changes. The foot straps that secure the foot to the base are made of thermoplastic polyurethane (TPU), a 3D-printed, flexible plastic. The compliance characteristic of TPU allows it to wrap around the user's shoe to maintain a firm attachment. Hook-and-loop fasteners allow the TPU straps to adapt to different shoe sizes and elastic bands attached to the heel strap allow for passive heel-lifting motion of up to 35 degrees, as shown in figure~\ref{fig:footmounting}. 
Toe-lifting motion cannot be achieved in the current design iteration. Having both passive heel and toe motion results in a very loose connection between the foot and platform, which was projected to have poor control performance, so only heel motion was chosen. The next iteration of the presented device will include increased functionality to address this limitation. Underneath the PLA base sits an ATI force sensor (Model MINI-58) to measure the interaction force, and a dampening material is implemented between the force sensor and foot platform to smooth the interaction forces for better pHRI stability~\cite{keemink2018admittance}. A support structure built of 40 mm by 40 mm T-slotted aluminum extrusions connects the gantry and the foot platform. The force sensor board housed in a 3D-printed case sits on the top of the support structure. 

\subsection{Power Delivery} 
\begin{figure}[b!]
\centering
\includegraphics[width=0.95\textwidth]{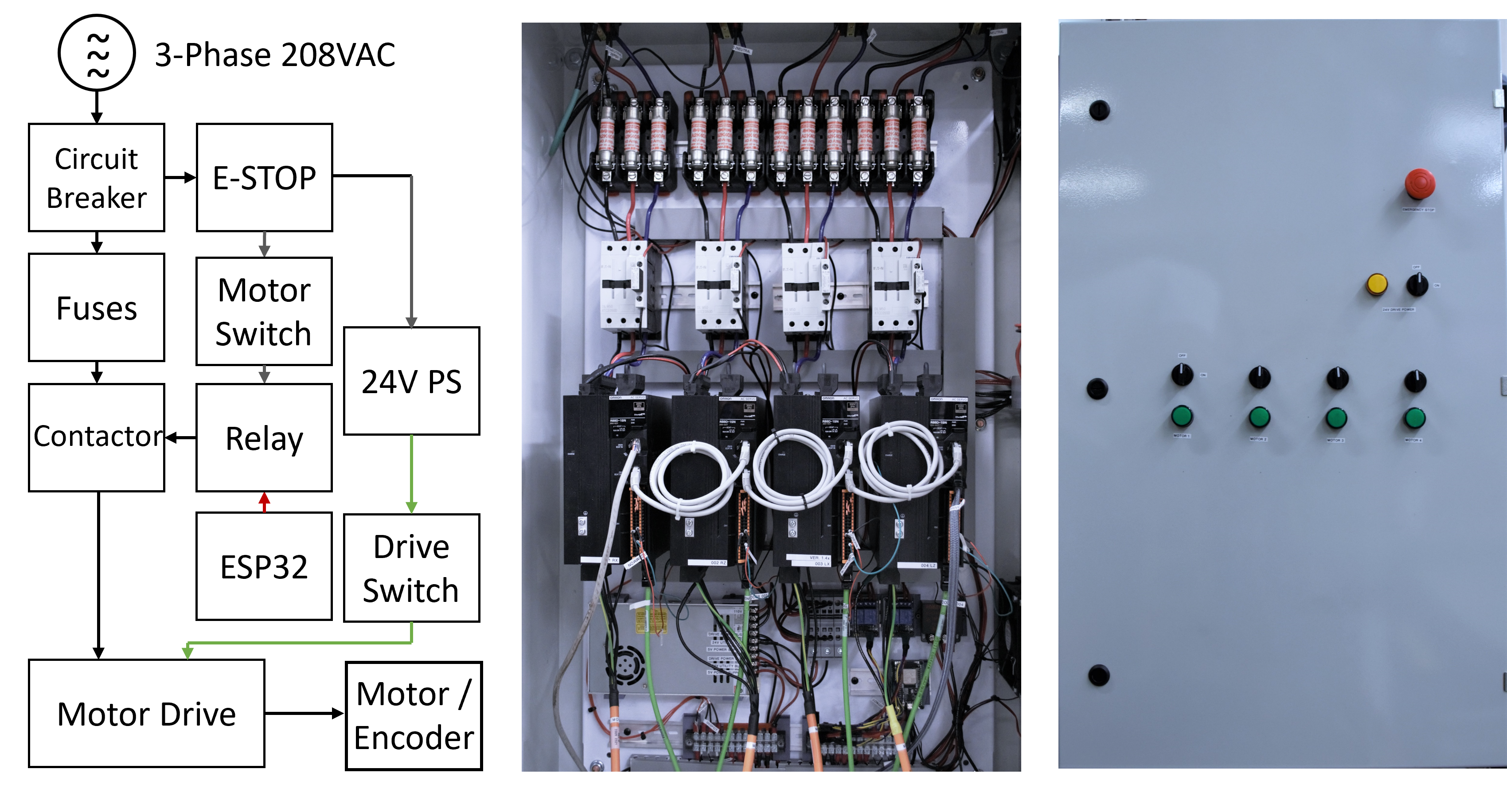}
\caption{Left: Simplified block diagram of the power delivery system for one motor unit. The 208VAC high power line and 24 VDC power line are colored black and green, respectively. The red line is the signal input to control the E-STOP relay. Middle: Interior of the power delivery system. From top to bottom are fuses, contactors, motor drives, and a 24V power supply (left). The circuit breakers are installed next to the power outlet, which is not shown in this. Right: On-site operation panel of the power delivery system. From the top to bottom are an emergency stop (red), a motor drive switch (yellow), and motor power switches (green).}
\label{fig:controlBox}
\end{figure}
The custom specifications of this locomotion interface also require a customized electric power delivery solution. This power delivery system is built to contain all the electronic components, including motor drivers, low-voltage power supplies, and an on-site operation panel. This system provides power to all the electrical components and controls the actuators. It receives commands from a central controller computer through the EtherCAT protocol. A wireless E-stop receiver is integrated into the system for experiment safety. Figure \ref{fig:controlBox} illustrates a simplified component diagram along with the exterior and interior of the power delivery system. The high-power 208VAC lines supply power to the motors, and the 24VDC low-power lines supply power to the motor drivers and other electrical components. This power distribution system consists of four motor drivers (Model R88D-1SN20H-ECT by Omron), circuit breakers (Model QO330 by Square D), fuses (Model A2K40R by MERSEN), contactors (Model XTCE050DS1E by Eaton), surge suppressors (Model XTCEXRSFB by Eaton), and a wireless emergency stop controlled by two ESP32 with 10Hz heartbeat signal. The wireless emergency stop is triggered manually or automatically when the connection between the transmitter and receiver is lost. Switches and LED indicators are installed in the front panel for on-site operation. All the electrical components used to build the power distribution hub are Underwriter Laboratories (UL) listed and used according to manufacturer recommendations for safe power distribution.

\section{System Framework}\label{sec3}
\begin{figure}[b!]
\centering
\includegraphics[width=0.83\textwidth]{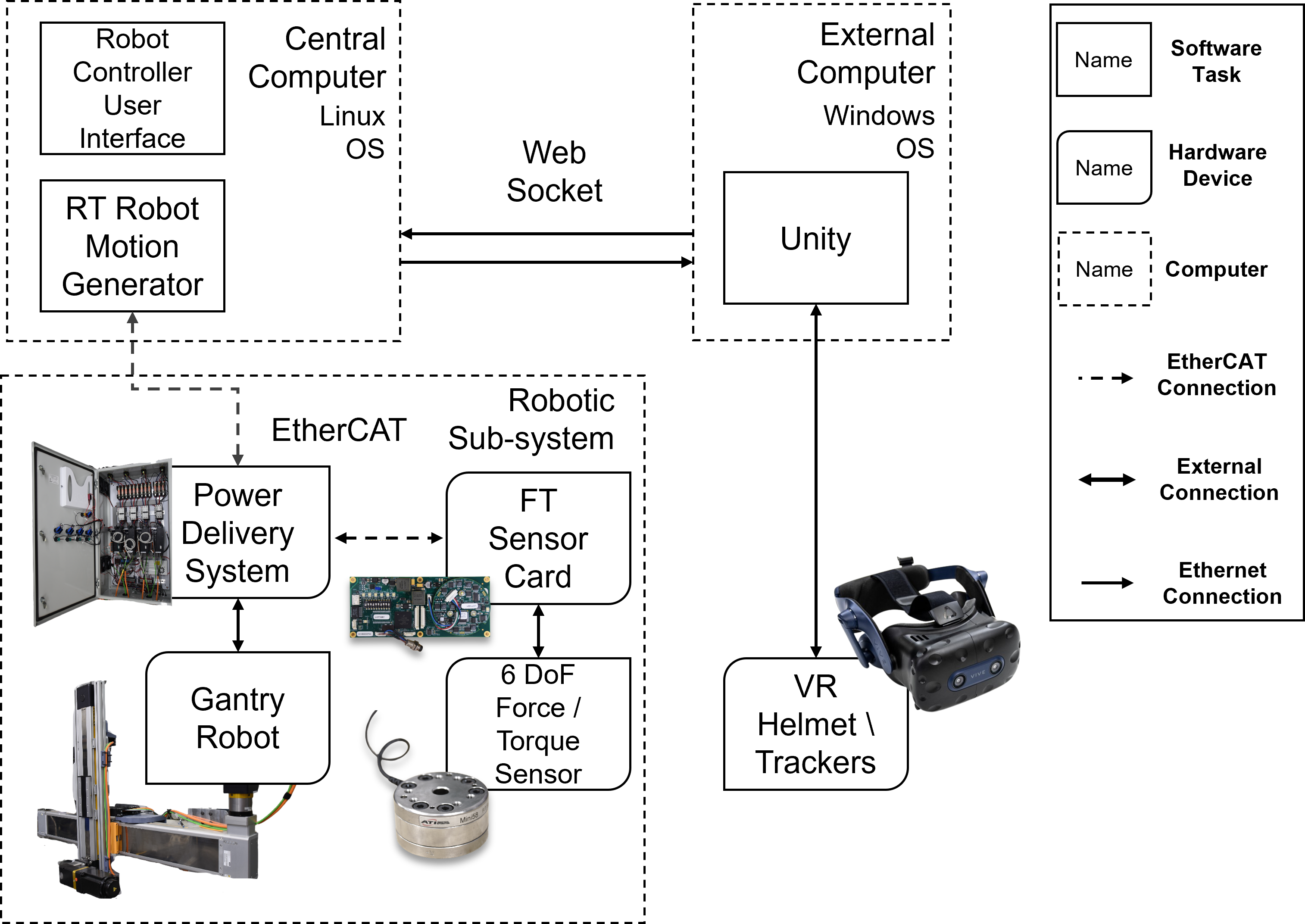}
\caption{System framework block diagram that made of three segment: (i) a central computer, (ii) the robotic sub-systems, and (iii) an external computer. The central computer connects to the power delivery system via EtherCAT to control the motors and to receive sensor information. The central computer also connects to the external computer via web sockets. The external computer runs the VR engine in conjunction with the connection to the VR equipment.}
\label{fig:systemArchetecture}
\end{figure}

The system framework illustrates the physical layer connections linking the robotic system with a central controller computer. This system framework is categorized into three segments, shown in Fig.~\ref{fig:systemArchetecture}: (i) a central computer, (ii) the robotic sub-systems, and (iii) an external computer. This framework yields several advantages, including its scalability in adding more devices and support for EtherCAT real-time communication. This framework grants accessibility to the ease of upgrading in each segment to improve computation rate, which is crucial for haptic interfaces, as underlined by previous research~\cite{colgate1994, o2009improved}.  
The central computer serves as the core of the locomotion interface system, computing the high-level control task to regulate pHRI. It establishes communication links with all robotic subsystems and an external computer through EtherCAT and Websocket. This central unit is an Intel NUC mini PC featuring an 11th Generation i7-1165G7 processor running on the Linux operating system. It runs an open-source robotic software provided by the Institute for Human and Machine Cognition (IHMC), incorporated with the Simple Open EtherCAT Master (SOEM) to be an EtherCAT master. Both the IHMC software control task and EtherCAT protocol run at 1,000 Hz on soft real-time, providing a sufficient computation rate for pHRI. The robotic sub-systems include the force sensors, power delivery and the gantry systems. It sends the sensor information to the main computer and executes the low-level control commands through EtherCAT. The ATI force sensor operates on a CANBus protocol, which is commonly used in hardware communication. To bridge between the EtherCAT and CANBus protocol, an ESP32 is paired with an EasyCAT Pro shield (AB\& T) to join the force reading into the EtherCAT network. 

The third component is the external computer, which runs the VR rendering software to give the user visual feedback from the VR world through a VR headset. The external computer permits the integration of different operating systems. It simplifies the development of establishing connections with commercially available VR products that are not compatible with Linux OS on the central computer. Communication between the external computer and the central computer is established via the Web-Socket protocol. This protocol, characterized by an HTTP-based, full-duplex mechanism, enables simultaneous data transmission and reception. This bidirectional protocol establishes a soft real-time communication channel that bridges different computers. Previous work has used this protocol for communication with the VR software Unity~\cite{ben2022}, but the framework can be used with any VR software. The presented framework is a generic haptic interface development solution, offering a 1,000 Hz soft real-time computation rate with EtherCAT communication. It provides accessibility to commercially available VR products such as VR helmets, motion trackers, haptic suits, and haptic gloves. Developers can integrate these VR devices for different purposes, providing a versatile solution to save development costs.

\section{Haptic Interface Controller}\label{sec4}
\begin{figure}[b!]
\centering
\includegraphics[width=0.5\textwidth]{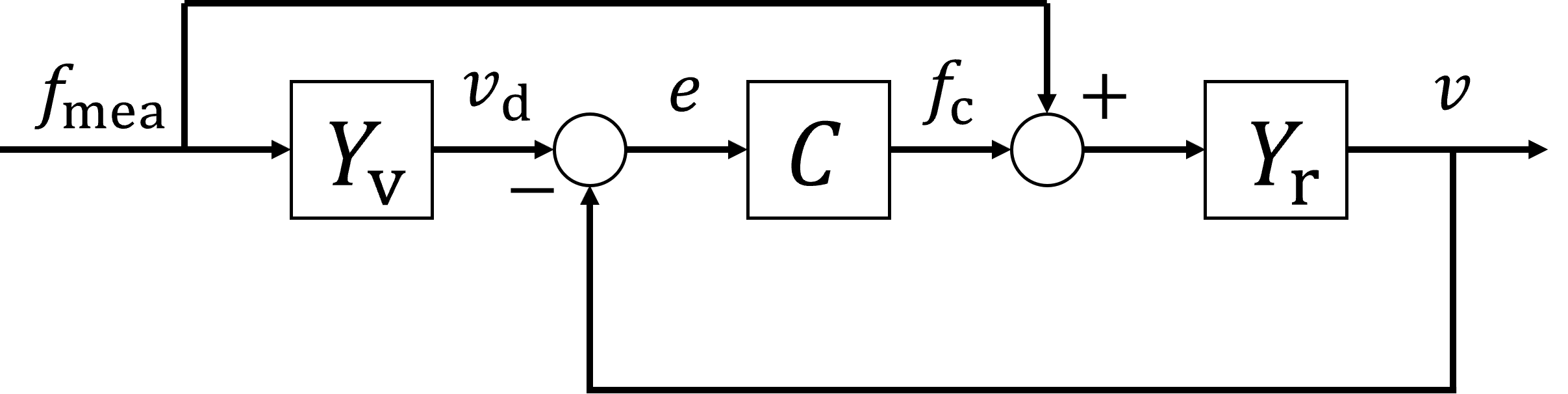}
\caption{A simplified block diagram of an admittance-control-based device. The measured interaction force denoted as $f_{\rm mea}$ is fed into an admittance controller $Y_{\rm v}$, yields the desired velocity $v_{\d}$. This desired velocity is governed by a low-level controller $C$, which applies mechanical force $f_{\rm c}$ to a robotic system $Y_{\rm r}$. The error $e$ is the difference between the desired velocity $v_{\d}$ and the actual velocity $v$.}
\label{fig:blockDia}
\end{figure}

For haptic interfaces, two common control strategies for regulating pHRI are to control the end-effector's impedance or admittance~\cite{kurfess2005robotics}, the former "measures motion and displays force," and the latter "measures force and displays motion"~\cite{yoshikawa1995}. The impedance controller regulates the dynamic interaction between a robot and a user by controlling the mechanical impedance of a haptic interface while tracking a desired motion defined by user input. Thus, it can be expected that the device will need to be back-drivable or have a zero force/torque controller implemented on the joint level to allow motion to happen before applying force to the user. Unlike impedance control, admittance control regulates the motion of a robot based on desired position and velocity commands derived from user input force (fig.\ref{fig:blockDia}). The choice between impedance or admittance control depends on the pHRI that developers want and the hardware they are using. Impedance control is often preferred for rendering low inertia, while admittance control is favored for rendering stiff surfaces but not low inertia~\cite{keemink2018admittance}. An admittance controller is implemented in this work for its advantage of displaying a stiff surface with the non-back-drivability inherent in the system. 

A locomotion device has two primary tasks for different gait phases: displaying a virtual ground in the stance phase or rendering low inertia (free-motion) during the swing phase. The presented work implements an admittance controller to achieve both task objectives. The admittance controller changes the output velocity of the foot platform to regulate the interaction force with the user, as shown in figure~\ref{fig:blockDia}. Since the actuators in this device are highly non-backdrivable, an admittance controller can be directly implemented without further modification compared to an impedance controller. The admittance controller generates velocity commands to follow the user's motion according to the input forces during the swing phase and tunable virtual mass and damping parameters. During stance phase, the desired velocity on the Z-axis is directly commanded to be zero to simulate a virtual ground.

The admittance controller can be described as 
\begin{align}
    f_{\rm mea} & = m\v \dot{v}\d + c\v v\d~, \label{eq:admittanceBaseEq}
\end{align}
where $m_{\rm v}$ and $~c_{\rm v}$ represent the virtual mass and virtual damping, $f_{\rm mea}$ is the measured interaction force. The admittance controller  $Y_{\rm v}$ acts as a reference model and generates the desired velocity trajectory $v_{\rm d}$ subject to the interaction force $f_{\rm mea}$. The controller transfer function can be derived from the Laplace transform of equation~\ref{eq:admittanceBaseEq}\begin{align}
    F_{\rm{mea}}(s) &= m_{\v} s V_{\d}(s) + c_{\v}V_{\d}(s)\\
    \frac{V_{\d}(s)}{F_{\rm{mea}}(s)} &= \frac{1}{m_{\v} s+ c_{\v}}\\ 
    Y_{\v}(s) &= \frac{\nicefrac{1}{c_{\v}}}{\frac{m_{\v}}{c_{\v}}s+1}~, \label{eq:admTrans}
\end{align} where $F_{\rm mea}(s)$ denotes the measured interaction force and $V_{\d}(s)$ is the desired velocity in the s-domain. Lastly, $Y_{\v}$ is the admittance, a ratio of robot output motion to the user input force. In equation~\ref{eq:admTrans}, we can see how the virtual damping $c_{\v}$ directly affects the magnitude of the output response, and the ratio of virtual mass and damping $\frac{m_{\v}}{c_{\v}}$ determines the response time. Multiplying the measured force by eq.\ref{eq:vdTrans} results in the desired velocity \begin{align}
    V_{\d} (s) &= F_{\rm mea}(s)Y_{\v}(s) =\frac{\nicefrac{F_{\rm mea}(s)}{c_{\v}}}{\frac{m_{\v}}{c_{\v}}s+1}~, \label{eq:vdTrans}
\end{align} which can be implemented as a discrete-time system  \begin{align}
    v_{\d}(k) &= \frac{f_{\rm{mea}}(k)-c_{\v} v_{\d} (k-1)}{m_{\v}}T_{\rm{s}}+v_{\d} (k-1)~, \label{eq:vdDiscet}
\end{align} where $T_{\rm{s}}$ is the sampling time period, and $k$ being the time step.

A system is at low admittance when virtual mass and damping are set to a high value, requiring a larger input force from actuators to achieve a desired speed or acceleration. A general understanding of the admittance controller is high admittance for large and fast pHRI movement and, conversely, low admittance for precise movement~\cite{Lecours2012}. In this work, the admittance controller's virtual mass and damping values are experimentally tuned to be as low as possible to minimize the user input force during swing while ensuring stability. Instability here is referred to oscillatory motions that the user can perceive during pHRI.

\section{Walking algorithm}\label{sec5}
\begin{figure}[h]
\centering
\includegraphics[width=0.6\textwidth]{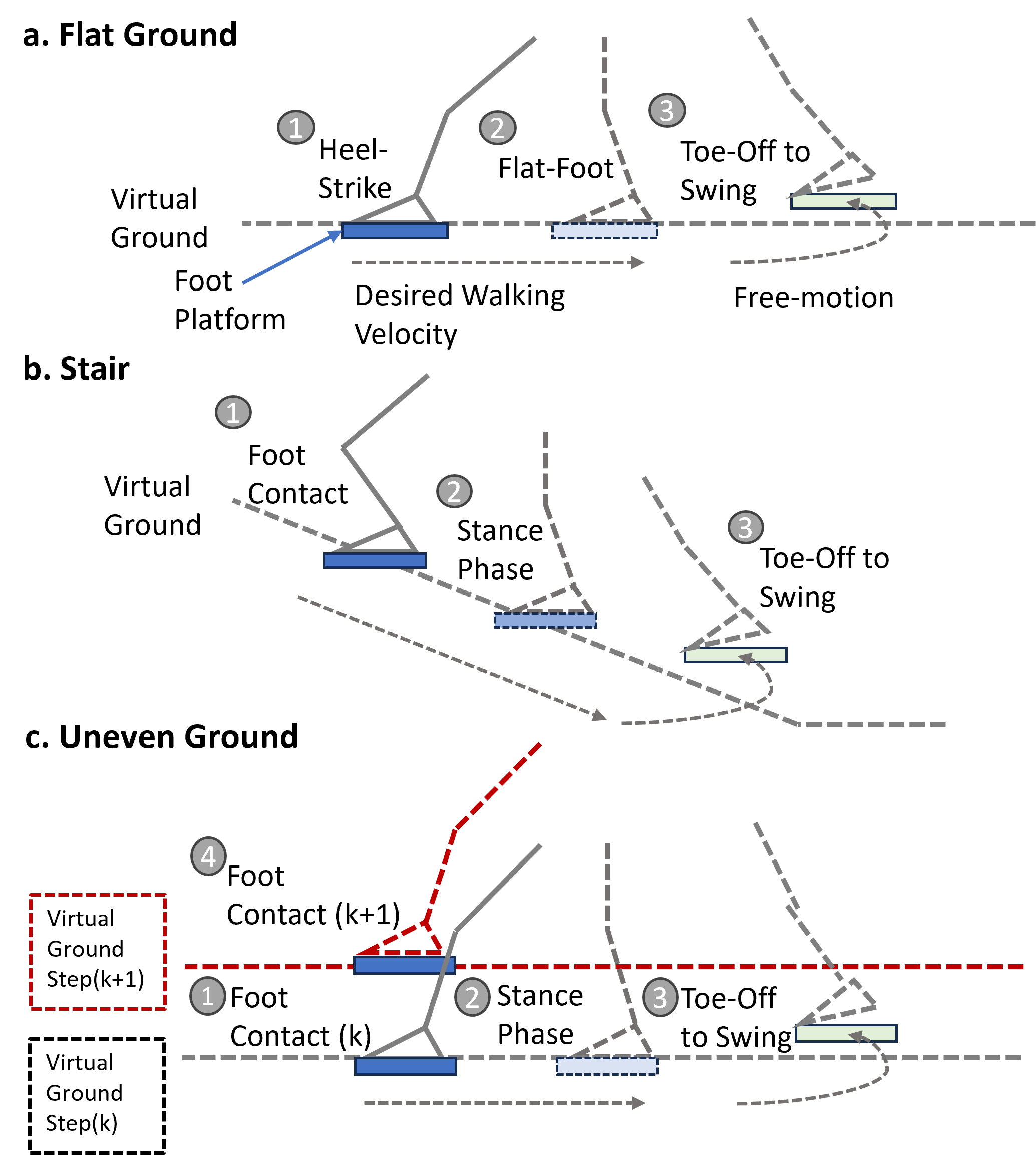}
\caption{An illustration of how the walking algorithm delivers walking motion. The foot platform slides with a pre-determined desired walking speed in the stance phase, transitioning to free motion upon the toe-off and the swing phases with an admittance controller. Case a: Flat ground, the foot platform slides along a flat virtual flat surface. Case b: Stair, the platform slides along a virtual slope. Case c: Uneven ground, the virtual ground height changes during each step.}
\label{fig:walkingAlg}
\end{figure}
This walking algorithm generates robotic motion according to gait phases to provide walking interaction. The gait phases are divided into the swing and stance phases depending on the foot platform position. If the foot platform position is in contact with a virtual ground, it is classified as the stance phase. Otherwise, the gait will be classified as the swing phase if it is above the virtual ground. Users can disengage the stance phase by applying an upward Z-axis force, resulting in a stance-to-swing phase transition. This position-based gait classification is applied to each foot individually and does not explicitly address the double stance phase, where both feet are on the ground. During the stance phase, the foot platform supports the user's weight by overwriting the admittance controller. A zero desired velocity is commanded on the Z-axis, which simulates a simple virtual ground. At the same time, the foot platform is moved along the virtual surface on the X-axis to emulate a treadmill and provide a walking experience to the user as figure~\ref{fig:walkingAlg}-a shows. Altering the speed at which the stance foot is moved changes how fast the user feels that they are moving. For the swing phase, the foot platform follows the user's foot movement using the admittance controller, letting the user lead the robotic foot platform. 
\begin{figure}[b]
\centering
\includegraphics[width=0.6\textwidth]{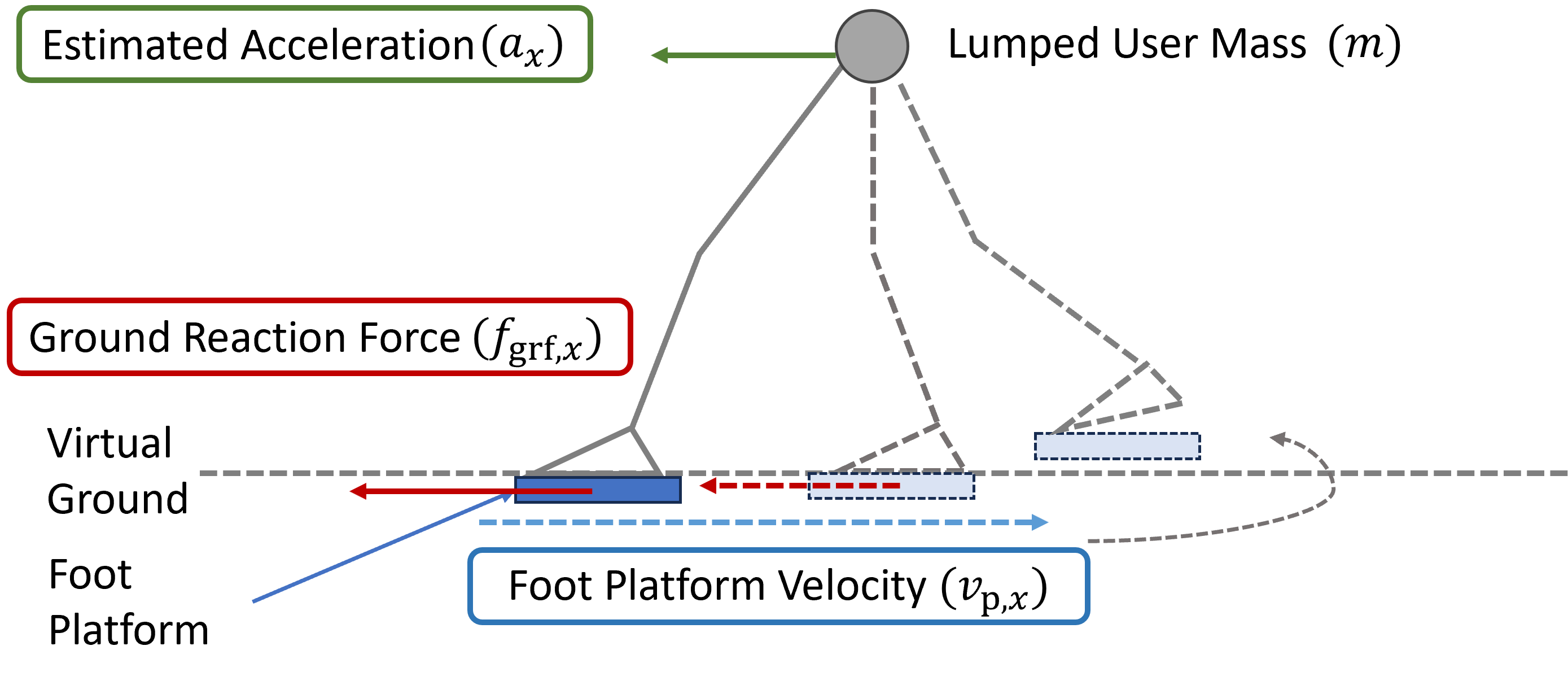}
\caption{Dynamic walking velocity estimation using ground reaction forces and lumped user mass.}
\label{fig:dynWalking}
\end{figure}
The virtual ground can be configured to represent various virtual terrains. Here, we included two example cases as a demonstration in figure~\ref{fig:walkingAlg}-b,c. The first example is a staircase (fig.\ref{fig:walkingAlg}-b), where the virtual ground is configured as a slope. Instead of sliding the platform along the X-axis like a treadmill in the previous case, the foot platform now slides along a virtual slope, moving in both the X and Z-axis. Due to the limitation of an unactuated ankle joint, the virtual slope can not be presented in this case. And thus, the interaction feels more similar to walking on an escalator than an actual slope. The third example is an uneven terrain case, demonstrating how to represent different VR terrain heights. In this case, the ground height changes at each step without vertical compensation as in the second case. In the VR application, these ground heights can be adjusted according to virtual reality ground height to reflect the virtual terrain. 

\subsection{Dynamic Walking Velocity}\label{sec5.1}
Dynamic walking speed control on a locomotion interface can be achieved by estimating the user's intended walking velocity using biomechanical measurements. Key indicators for this estimation include gait period~\cite{Yoon2010a}, swing foot velocity~\cite{Yoon2009}, and reference positions~\cite{Kim2023}. Ground reaction forces during push-off, which show a positive correlation with walking speed~\cite{Dong2010, Yu2021}, and the center of pressure (COP)~\cite{Feasel2011}, also provide valuable cues for speed adjustment. These methods rely on indirect estimations of user velocity derived from biomechanical correlations. However, such correlations often exhibit significant variability across individuals.

In order to achieve a realistic VR interaction, we introduce a walking speed estimation based on a single rigid body assumption, using a dynamic approach to enable the system to adapt to the user's intended walking velocity. The estimation method uses a lumped user mass ($m$) and ground reaction force along the $x$-axis ($f_{\text{grf},x}$) to calculate the acceleration ($a_x$) \begin{align}
    a_x(k) &= \frac{f_{\text{grf},x}(k)}{m}.
\end{align} The estimated user velocity ($v_x$) can be obtained by integrating the estimated acceleration over the computation time period from timestep $k$ to $k+1$ \begin{align}
    v_{x}(k+1) &= v_{x}(k) + a_x(k)T_\text{s},
\end{align} and the foot platform in $x$-axis ($v_{\text{p},x}$) is equal to the negative estimated velocity to compensate for walking motion~\begin{align}
    v_{\text{p},x}(k+1) &= -v_{x}(k+1),\label{eq:walkingSpeed}
\end{align}~as shown in figure~\ref{fig:dynWalking}. 

\subsection{Vertical Movement Compensation}\label{sec5.2}
\begin{figure}[h]
\centering
\includegraphics[width=0.55\textwidth]{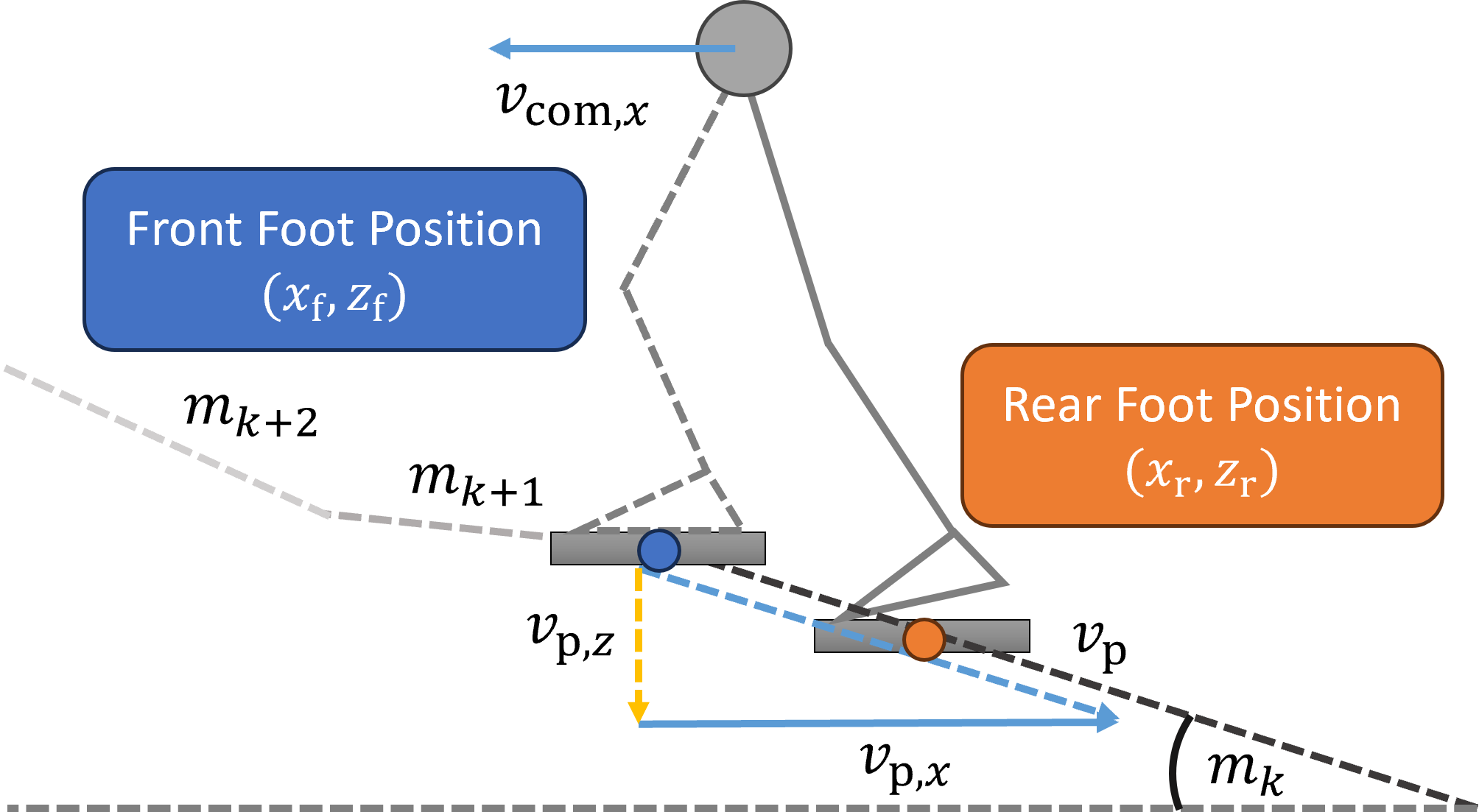}
\caption{Vertical Motion Compensation. The algorithm records the foot position at each heel strike to calculate the slope between the forefoot and rearfoot. Using this slope and the estimated walking velocity, the foot platform's vertical velocity along the z-axis is determined, enabling dynamic adjustment for slope traversal.}
\label{fig:vertical}
\end{figure}

The walking algorithm introduced in section~\ref{sec5.1} generates robotic motion to compensate for forward movement, functioning similarly to a treadmill. In this section, we propose an approach that dynamically address scenarios where users traverse terrains with variable heights, compensating vertical movement by leveraging the estimated walking velocity and the virtual terrain height on each step, as shown in Figure~\ref{fig:vertical}.

ForceBot logs foot positions whenever a heel-strike event is detected and uses this data to compute the slope of the terrain the user is navigating. Once a heel-strike event is identified, the walking algorithm classifies the forefoot and rearfoot to determine the virtual slope. While traversing the virtual slope, the foot platforms compensate for motions along both the $x$- and $z$-axes. The $x$-axis motion ($v_{p,x}$) moves backward which is equal to the negative of the estimated walking velocity ($-v_x$), as defined by equation~\ref{eq:walkingSpeed}. While the $z$-axis velocity ($v_{\text{p},z}$) is calculated by \begin{equation}
    v_{\text{p},z} = m_k \cdot v_{\text{p},x},
\end{equation} where $m_k$ is the slope calculated by the relative position between the forefoot and rear foot on each footstep $k$ \begin{equation}
    m_k = \frac{z_\text{f} - z_\text{r}}{x_\text{f} - x_\text{r}}.
\end{equation} The forward ($v_{\text{p},x}$) and vertical ($v_{\text{p},z}$) velocities work together to bring the foot platforms backward, and downward, as if the user is walking on an escalator. Lastly, the total travel distance ($d_x$) and height ($d_z$) can be calculated as the integration of the foot platform velocities~\begin{align}
    d_x(k+1)&= d_x(k) + v_{p,x}(k)T_s,\label{eq:xdist}\\
    d_z(k+1)&= d_z(k) + v_{p,z}(k)T_s.\label{eq:zdist}
\end{align} This estimated travel translates the user's foot movements into the virtual avatar's body position. This follows exactly the same principle as the treadmill, in which the total distance a user travels over the treadmill is equal to the belt speed over a given time period.

\subsection{Virtual Reality Integration}\label{sec5.3}
\begin{figure}[h]
\centering
\includegraphics[width=1\textwidth]{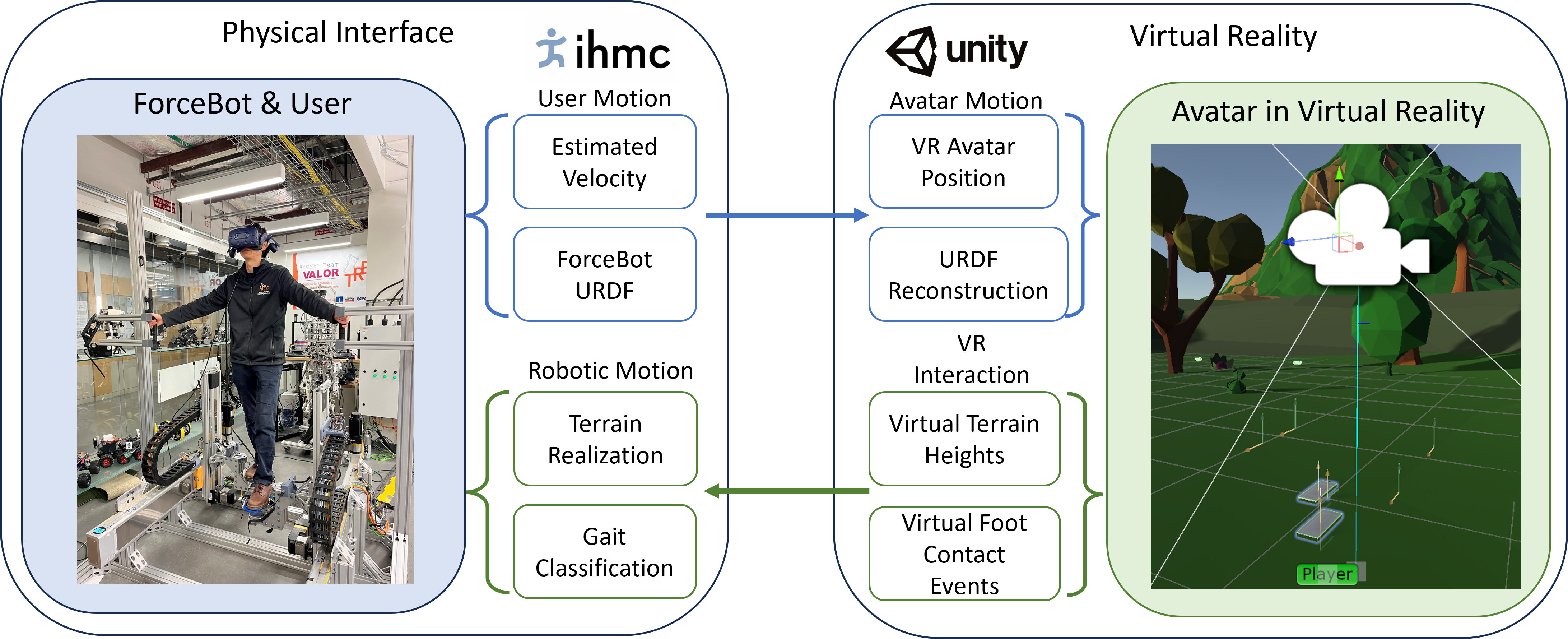}
\caption{The system framework for VR terrain realization. It contains two parts: Left: the physical interface shown in blue blocks. Right: the VR part contains Unity, and the virtual environment configuration is in green blocks. The left part of the figure contains the ForceBot locomotion interface and the underlying algorithms run in the IHMC open robotic software to estimate user-intended walking velocity. The estimated walking velocity is then sent to Unity for VR avatar reconstruction, translating the user's foot movement into VR avatar body position. ForceBot then realizes virtual terrains based on the interaction that happens in the VR.}
\label{fig:vrFramework}
\end{figure}

The previous section describes the framework of a walking algorithm, which uses swing and stance phases to switch between different controllers, providing ground support or free swing motion and calculating virtual slopes. Classifying the swing and stance phases depends on the foot interaction in VR. This subsection presents a hardware and software framework that allows ForceBot to communicate with Unity to realize VR interaction. This bidirectional communication sends user motion to VR while retrieving information from Unity. Firstly, the estimated walking velocity is translated into VR avatar body position using eq.~\ref{eq:xdist} and eq.~\ref{eq:zdist}. At the same time, ForceBot streams the foot platform position to Unity using URDF (unified robotics description format) to reconstruct the virtual avatar's foot positions. On the Unity side, two crucial pieces of information need to be returned from VR. The first is the ground contact event, where the VR avatar's foot is in contact with the ground. The second one is the virtual terrain height, which calculates the VR slope and realizes the virtual terrain, as shown in figure~\ref{fig:vrFramework}. Under this framework, the physical interface does not have any knowledge of how the VR world is configured; it solely relies on the retuning information from Unity to realize VR interactions. 

\section{Experimental Validation \& VR Interaction}\label{sec6}
\begin{figure}[t!]
\centering
\includegraphics[width=0.6\textwidth]{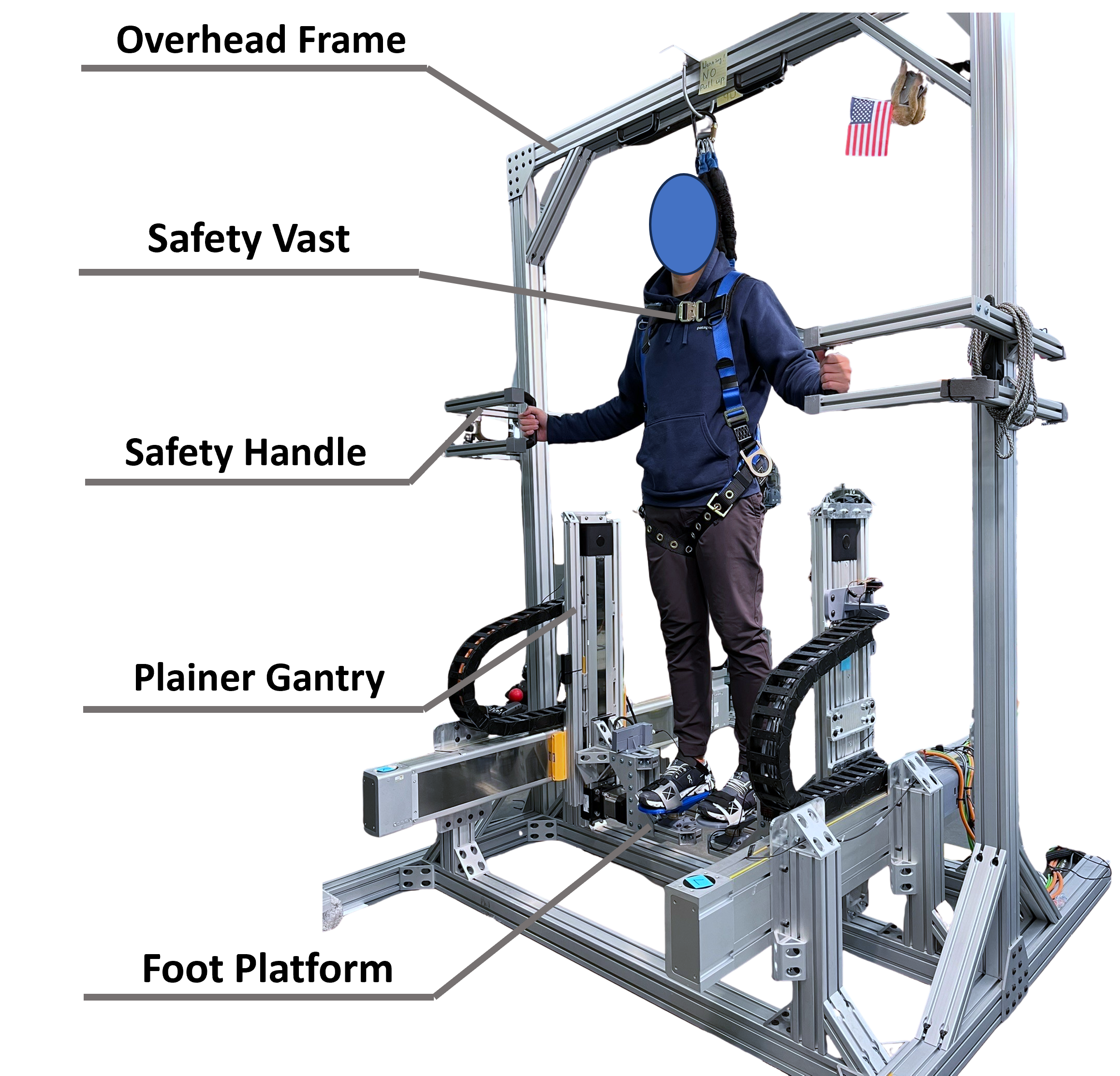}
\caption{The experiment setup of the locomotion device. An overhead frame is connected to the safety vest for fall arrest and unbalance. The user's feet are attached to the foot platforms to interact with the device. Lastly, a pair of safety handles is provided for the user to grip.}
\label{fig:expSetup}
\end{figure} 
This section contains two sections to validate the presented approach for realizing VR lower-body interaction. The first part evaluates the system's fundamental performance, such as the delay or force-to-velocity relationship. This part does not contain VR integration and stays in a fixed walking velocity for repeatability. The second part focuses on the application of VR interaction with dynamic walking velocity estimation to showcase the device's capability.

\subsection{Performance Evaluation}
The first validation section examines the system's performance through three experiment measurements. The pilot test subject interacts with the interface and performs walking tasks with a constant velocity of 400 mm/s. The foot platform trajectory, velocity, interaction force, and other data are logged during the experiment for analysis. Since both feet have an almost identical walking pattern, only the right foot data is presented in the result section. The admittance controller is tuned experimentally to have $m\v = 8$ kg of virtual mass and $c\v = 4$ Ns/m virtual damping, maintaining pHRI stability while minimizing input force. The experiment setup is shown in figure~\ref{fig:expSetup}, having the user attach their feet to the locomotion device and wearing a fall arrest vest for safety. A pair of handles is provided for the user to grip during experiments for fall prevention. The user is told not to grip the handles firmly in case it disrupts the gait. The presented work has taken inspiration from previous locomotion interface development studies, but there is a need for a more detailed evaluation of user data or system dynamic performance. Previous studies either do not include preliminary user data~\cite{schmidt2005hapticwalker,Iwata20010} or the user being constrained in a small workspace with a slow walking gait~\cite{yoon2010,yoon2010planar}. This study presents the following experiment data to contribute to the locomotion interface community for future device development. 
\begin{figure}[b]
\centering
\includegraphics[width=1\textwidth]{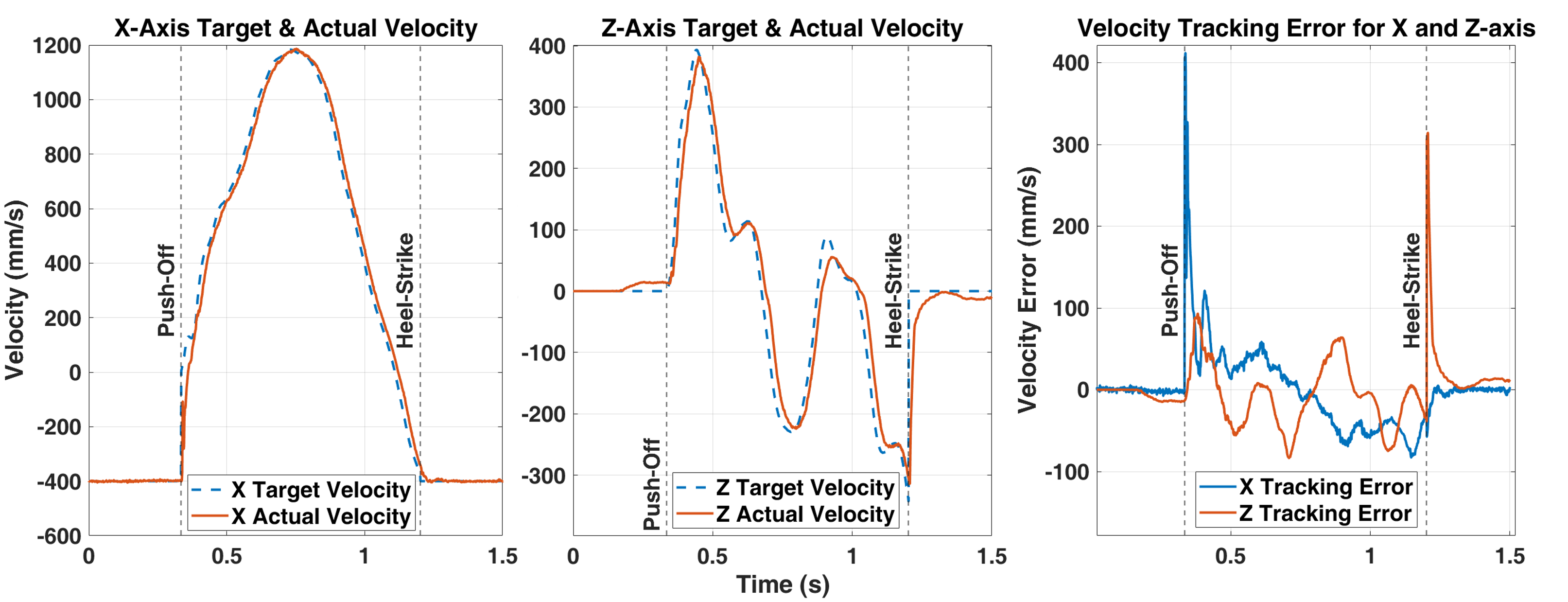}
\caption{The velocity tracking performance of the X-Axis (left) and Z-Axis (center) actuators. Exact tracking error is shown (right) with stance-to-swing and swing-to-stance transitions happening at 0.33 sec and 1.2 sec. 
}
\label{fig:xzvel}
\end{figure}

\begin{figure}[t]
\centering
\includegraphics[width=0.4\textwidth]{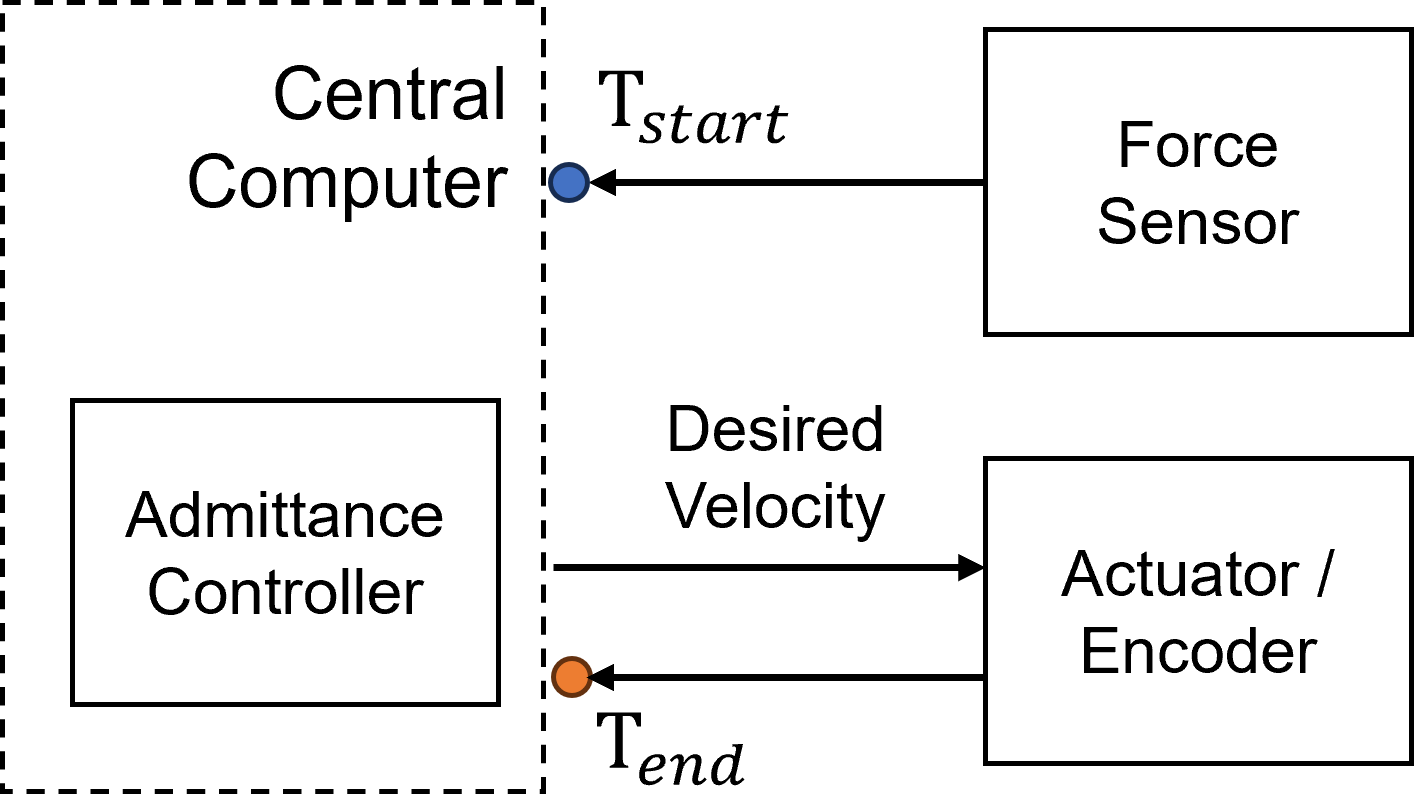}
\caption{The velocity profile delay measures a period from receiving a force reading to an actual robotic motion being recorded, from time $T_{\rm{start}}$ to time $T_{\rm{end}}$.}
\label{fig:meaDelay}
\end{figure}

The first measurement to evaluate the performance of the system is the velocity tracking performance, comparing the target velocity and actual velocity during the swing phase, as shown in figure~\ref{fig:xzvel}.  Other than high tracking error spikes during stance and swing state transitions (at about 0.33 s and 1.2 s) due to large instantaneous desired velocity changes, the tracking error mostly stays well within 100 $\nicefrac{mm}{s}$. The velocity profile tracking delay ranges from 6 ms to 20 ms depending on the given target velocity, with a larger delay of up to 100 ms at state transitions, where an instantaneous change in desired velocity is commanded. Previous studies rarely include delay measurement, except a 300 ms sensor delay in~\cite{Iwata20010}, or a 200 ms position control delay in~\cite{yoon2010planar}. Part of this delay is due to a communication and processing delay measured at approximately 4 to 6 ms. This delay measures the time stamp difference from a force sensor signal reading ($T_{\rm{start}}$) to an actual robotic motion recorded ($T_{\rm{end}}$), shown in figure~\ref{fig:meaDelay}. The system is currently running at 1,000 Hz and it takes at least three communication cycles from receiving force signal to receiving encoder reading, so the measured delay sits in a reasonable range, since the system can drop a couple frames. This delay can be improved by having a higher computation and communication rate, which would reduce each communication cycle, ultimately leading to a more responsive pHRI.     

\begin{figure}[b]
\centering
\includegraphics[width=1\textwidth]{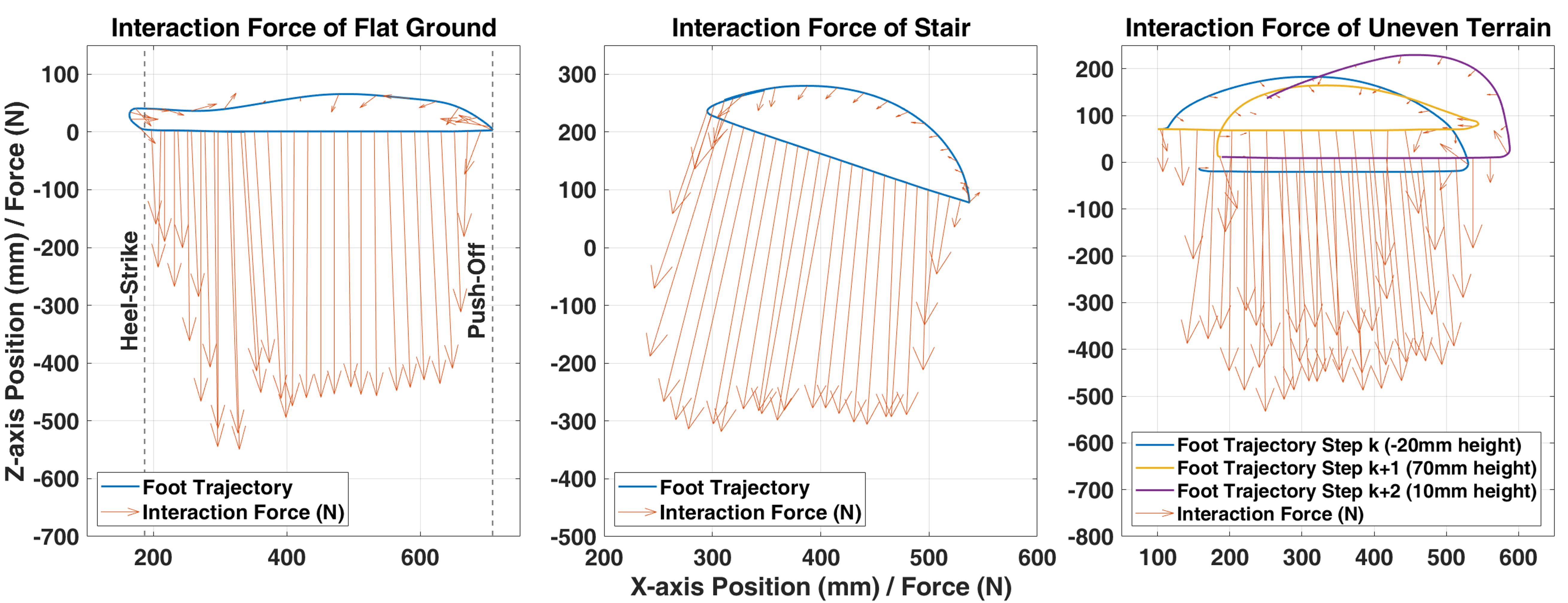}
\vspace{-10pt}
\caption{The interaction force and the gait trajectory when walking on the locomotion device under three different terrain simulations. Orange vectors represent the interaction force and direction along the foot platform trajectory. And the blue line is the foot platform trajectory. Left: gait trajectory versus interaction force in flat-ground simulation. The dashed lines indicates for the push-off and heel-strike. Middle: gait trajectory versus interaction force in stair simulation. Right: gait trajectory versus interaction force in uneven terrain simulation, three foot steps presented in the figure labeled in different colors.}
\label{fig:footTrajectory}
\end{figure} 

The second measurement is the foot platform position trajectory versus interaction force in simulating three different terrain geometries to validate the system's capability. Figure~\ref{fig:footTrajectory} shows the direction of the ground reaction and interaction forces at every point in a walking cycle. The walking velocity is fixed at 400 $\nicefrac{mm}{s}$ in the flat-ground case (fig.~\ref{fig:footTrajectory}-left), and is tuned down to 200 $\nicefrac{mm}{s}$ for in both stair(fig.~\ref{fig:footTrajectory}-middle) and uneven(fig.~\ref{fig:footTrajectory}-right) terrain, since users tend to walk slower among these two cases. The admittance controller successfully regulates the pHRI. It can be observed that the force is driving the foot platform during the swing motion. Upon heel-off, the forces point towards the walking direction and then reverse its direction at the mid-swing to slow down the foot platform. During the stance phase, that system supports the user's weight to simulate a virtual ground, resulting in interaction force pointing downward. This measurement justifies the system's capability to simulate simple terrain cases. The user is able to utilize twice as large of a workspace with much faster gait compared to previous devices~\cite{yoon2010planar}.

This measurement helps developers understand the relationship between the input force and the robotic motion, spotting unnatural interaction forces to improve the system. For example, fig.\ref{fig:footTrajectory} shows that the user exerts much force at the push-off phase to "pull" the platform away from the ground, instead of utilizing ground reaction force to "push" off the ground. In natural walking, the push-off motion requires a minimum effort that utilizes ground reaction force to swing the leg forward instead of actively pulling it up. This suggests the presented walking algorithm may require push-off assistance or further development to represent a more realistic walking interaction. 

\begin{figure}[t]
\centering
\includegraphics[width=1\textwidth]{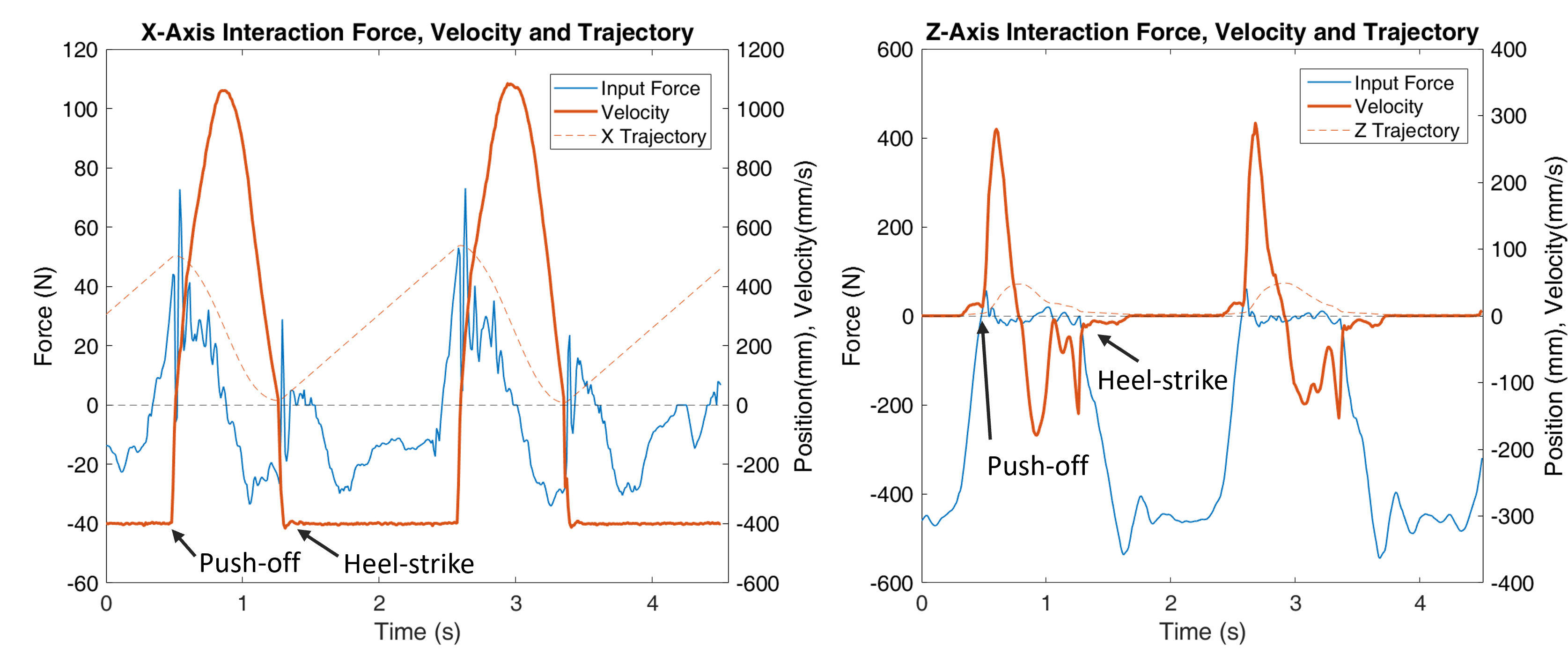}\vspace{-10pt}
\caption{The interaction force, foot platform velocity and position for X-axis (right), and Z-axis (left).}
\label{fig:xzvel_force}
\end{figure} 
The third measurement breaks down the interaction force into the X- and Z-axis to illustrate the relationship between force and velocity during flat-ground walking. The interaction force and the foot platform velocity are plotted in figure~\ref{fig:xzvel_force}, along with the foot platform position as a reference. The figure shows that the measured input forces that drive the pHRI mostly stay within 40 N during the swing phase despite having a transient peak force of 60 N. This result suggests that the pHRI is reasonably transparent and does not require extensive physical effort to operate the system. Experiment subjects also reported a similar conclusion that the experiment is not physically demanding and that they do not experience fatigue after a 5-minute experiment session. 

It can be observed that the interaction forces of both the X and Z axes oscillate upon push-off. Such oscillatory force can be a sign of potentially unstable pHRI. However, the user does not report perceivable oscillatory motion; this can be backed up by the fact that the X- and Z-axis position profiles are fairly smooth during the interaction. The oscillatory force is due to the robotic system dynamics, which requires time to settle to tracking large instantaneous changes in the user's desired motion. Lowering the admittance can suppress this oscillatory force but the trade-off is a higher interaction force, resulting in a less responsive pHRI interaction. This figure shows that the tuning of the virtual mass and damping parameters results in a well-balanced output performance. The user is easily able to perform walking interactions with the presented device.
\begin{figure}[h]
\centering
\includegraphics[width=0.6\textwidth]{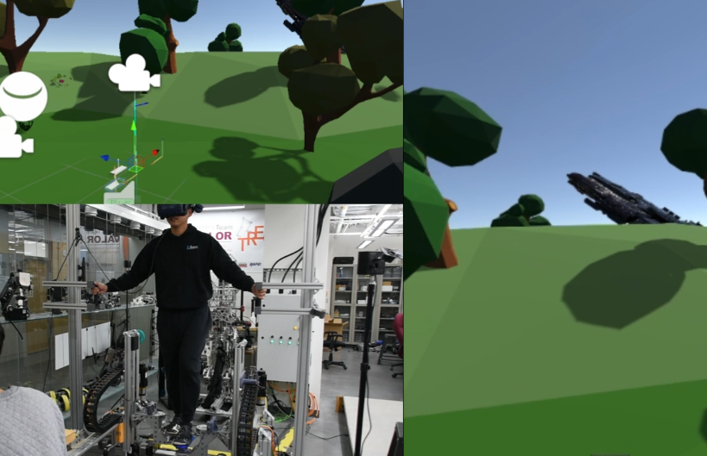}
\caption{The VR interaction setup with ForceBot and Unity. Top left: an overhead view of the VR world constructed in Unity. Bottom left: the user and experiment setup of the pHRI system. Right: the user's view provided by the VR helmet.}
\label{fig:vrSetup}
\end{figure}
\subsection{VR Interaction Result}\label{sec6.1}
The VR interaction section fully integrates the system with the VR, using Unity and the VR helmet to enable VR interaction. Furthermore, unlike the first validation section, which used fixed walking speeds to evaluate system performance, this section has implemented the dynamic walking velocity method from section~\ref{sec5.1} to estimate the user's walking speed, allowing the device to adapt to the user's intended walking velocity. Two experiment trials are conducted for evaluation: 1. a flat VR ground case to evaluate dynamic walking velocity estimation and 2. an up-hill climbing case to evaluate vertical movement compensation. Both experiments only differ in the VR terrain setup, and their underlying algorithm remains the same. Figure~\ref{fig:vrSetup} showcases the experiment setup with the device and the virtual world. The top left view in figure~\ref{fig:vrSetup} shows the VR avatar foot positions as the two white rectangular blocks. The bottom left view of fig.~\ref{fig:vrSetup} shows the entire system setup with a user on the device. Lastly, the right part of fig.~\ref{fig:vrSetup} shows the VR helmet view that the user is watching.
\begin{figure}[b!]
\centering\vspace{-10pt}
\includegraphics[width=1\textwidth]{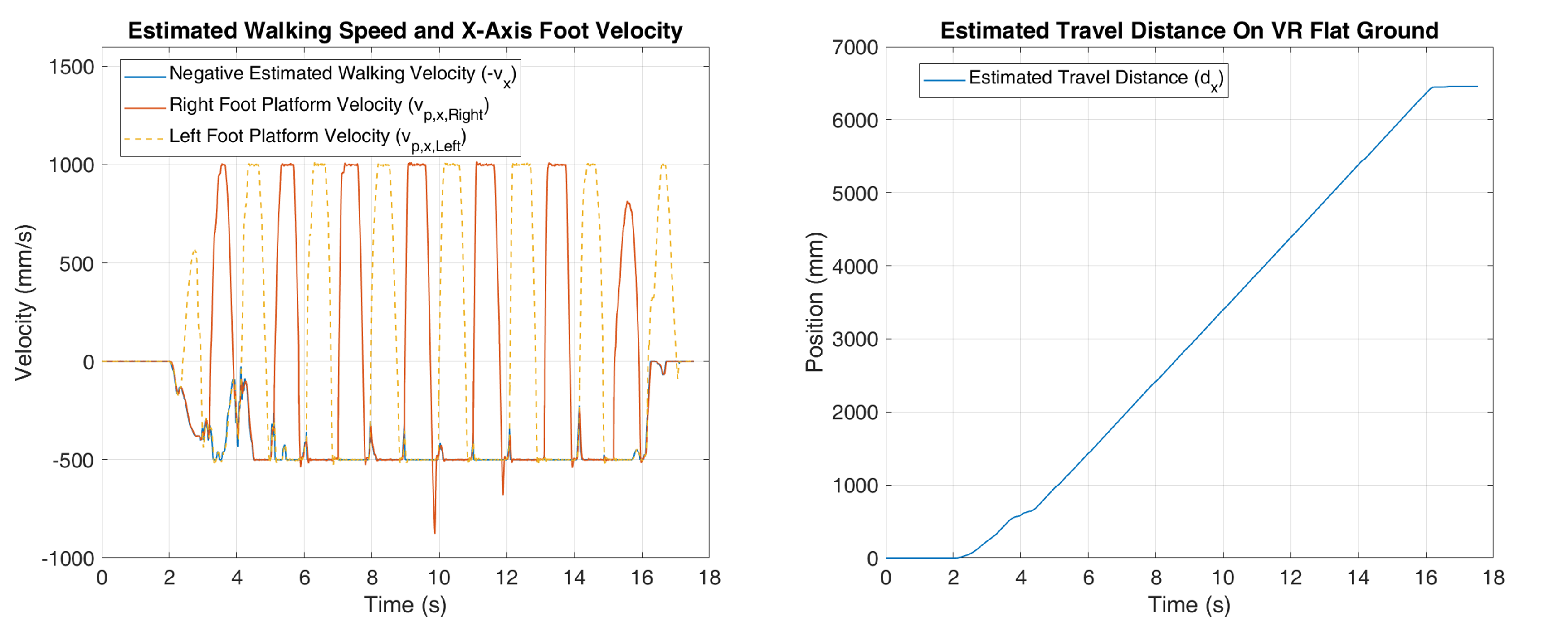}
\caption{The experiment result of a user walking over flat VR ground with dynamic walking speed estimation and travel distance estimation. Right: The relationship between estimated walking velocity ($v_x$) and the foot platform velocity. The estimated walking velocity replaced the admittance controller upon the heel strike. The estimated walking velocity is timed by $-1$ to show how it matches with the foot platform velocity during the stance phase. The right and left foot platform velocities are denoted by $v_\text{p,R}$ and $v_\text{p,L}$ respectively. Left: The estimated travel distance used for the VR avatar positioning.}
\label{fig:dynWalkingVel}
\end{figure}

Trial one, shown in figure~\ref{fig:dynWalkingVel}, successfully implements walking speed estimation over flat VR ground walking. For safety purposes, the maximum actuator speed is limited at 1,000 $\nicefrac{mm}{s}$, and the maximum walking speed is limited at 500 $\nicefrac{mm}{s}$. The estimated walking speed in figure~\ref{fig:dynWalkingVel} is set to be negative to showcase better how it matches the foot platform velocity during the stance phase, as presented in equation~\ref{eq:walkingSpeed}. The result shows that the user can initiate walking and then come to a stop at the end of the trail. At the same time, we can see each heel strike slightly affects the walking velocity due to the ground reaction force. The right part of fig.\ref{fig:dynWalkingVel} shows the estimated travel distance ($d_x$) using eq.~\ref{eq:xdist}.

\begin{figure}[t]
\centering
\includegraphics[width=1\textwidth]{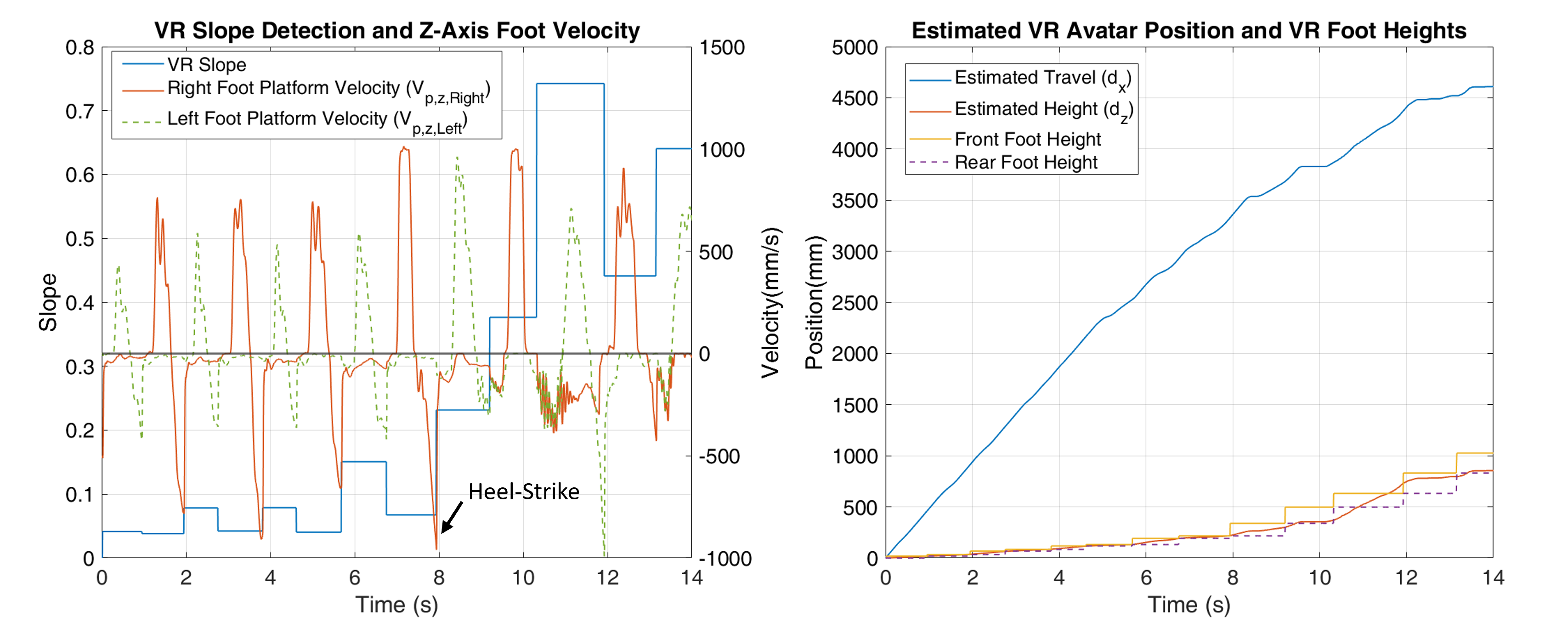}
\caption{Experiment data of a user traverse through VR up-hill terrain. Left: The VR slope detection and foot platform velocity on Z-Axis. Right: The estimated VR avatar travel and height with footstep heights.}
\label{fig:climbing}
\end{figure}

Trial two of the VR interaction experiment features a user climbing over an up-hill terrain, as shown in the top left of figure~\ref{fig:vrSetup}. This experiment focuses on validating the vertical movement compensation presented in section~\ref{sec5.2}. The result in figure~\ref{fig:climbing} demonstrates a successful implementation of vertical movement compensation, where the foot platform moves down and backward after the heel strike, as marked in the left part of fig.~\ref{fig:climbing}. It also shows the virtual slope detection on each step as the user traverses through the virtual terrain. The estimated travel height ($d_z$) nicely matches the VR footstep heights as shown in the right part of fig.~\ref{fig:climbing} using equation~\ref{eq:zdist}. The vibratory motion observed in fig.~\ref{fig:climbing} suggests further investigation is required for improvement. Overall, the presented results well demonstrated the capability of the presented device to realize virtual terrain, being one of the few devices in this field to achieve VR walking with realistic lower-body engagement.

\section{Conclusion}\label{sec8}
The presented work included a comprehensive development process of a robotic-platform-based locomotion interface. This document covers the locomotion system's design process, hardware, software, control, and experiment. This platform creates virtual terrain interaction, enabling users to navigate VE and experience the terrains as if they were in the virtual world. Building this platform allows for the analysis of lower-body virtual interaction along with the continued development and implementation of pHRI controllers. This work presents a unique design using two linear gantries that provide two DoF motions on each foot with a passive heel-lift mechanism to achieve walking behaviors. The design process utilized motion capture data with dynamic simulations to align the human-robot workspace with the equipment selection and hardware design built upon the simulation result. The design also provides a soft real-time framework with EtherCAT communication at a 1,000 Hz rate suitable for locomotion interface development. An admittance controller is used in the system for physical human-robot interaction that regulates the relationship between interaction force and robotic motion. This controller uses a walking algorithm to generate a treadmill-like walking motion. Lastly, the system's performance is characterized by three measurements that validate the system's capability to simulate virtual terrain.

Future works aim to develop and evaluate the presented device further. Firstly, we will focus on implementing more complex virtual environments, as well as further developing the walking algorithm. The improvements will include push-off assistance, providing more DoF on each foot, and simulating terrain properties, all of which will result in a more realistic walking experience. Secondly, future work will assess the effectiveness of the presented device through user studies. The study will involve a complete VR setting with inexperienced users for a comprehensive evaluation, including measuring the user's physical loading, VR-induced discomfort, and immersion level.

\section*{Acknowledgments}
This work is supported by NRI:INT:Collaborative Research (Award ID: 2024772) from the National Science Foundation (NSF). The authors would like to thank all the team members of the TREC Lab at Virginia Tech, especially, Michael Han, Suyeon Ahn, Melanie Hook, for their valuable work during the design and development of this project. 

\bibliographystyle{unsrt}  
\bibliography{main.bib}  

\begin{thebibliography}{10}

\bibitem{zhang2020}
Yuxuan Zhang, Hexu Liu, Shih-Chung Kang, and Mohamed Al-Hussein.
\newblock Virtual reality applications for the built environment: Research trends and opportunities.
\newblock {\em Automation in Construction}, 118:103311, 2020.

\bibitem{ali2017}
Nabeel Ali and Mohammed Al-Mhiqani.
\newblock Review of virtual reality trends (previous, current, and future directions), and their applications, technologies and technical issues.
\newblock {\em Journal of Engineering and Applied Sciences}, 12, 02 2017.

\bibitem{luis2020}
Luis Muñoz-Saavedra, Lourdes Miró-Amarante, and Manuel Domínguez-Morales.
\newblock Augmented and virtual reality evolution and future tendency.
\newblock {\em Applied Sciences 2020, Vol. 10, Page 322}, 10:322, 1 2020.

\bibitem{martin2018}
Jorge Martín-Gutiérrez, Carlos~Efrén Mora, Beatriz Añorbe-Díaz, and Antonio González-Marrero.
\newblock Virtual technologies trends in education.
\newblock {\em Eurasia Journal of Mathematics, Science and Technology Education}, 13:469--486, 1 2017.

\bibitem{app11062879}
Maged Soliman, Apostolos Pesyridis, Damon Dalaymani-Zad, Mohammed Gronfula, and Miltiadis Kourmpetis.
\newblock The application of virtual reality in engineering education.
\newblock {\em Applied Sciences}, 11(6), 2021.

\bibitem{sik2017virtual}
Cecilia Sik-Lanyi.
\newblock Virtual reality healthcare system could be a potential future of health consultations.
\newblock In {\em 2017 IEEE 30th Neumann Colloquium (NC)}, pages 000015--000020. IEEE, 2017.

\bibitem{Vorderer2021}
Peter Vorderer, Christoph Klimmt, Tilo Hartmann, and Jesse Fox.
\newblock {Entertainment in Virtual Reality and Beyond: The Influence of Embodiment, Co-Location, and Cognitive Distancing on Users' Entertainment Experience}.
\newblock {\em The Oxford Handbook of Entertainment Theory}, pages 717--732, feb 2021.

\bibitem{wang2018critical}
Peng Wang, Peng Wu, Jun Wang, Hung-Lin Chi, and Xiangyu Wang.
\newblock A critical review of the use of virtual reality in construction engineering education and training.
\newblock {\em International journal of environmental research and public health}, 15(6):1204, 2018.

\bibitem{berni2020applications}
Aurora Berni and Yuri Borgianni.
\newblock Applications of virtual reality in engineering and product design: Why, what, how, when and where.
\newblock {\em Electronics}, 9(7):1064, 2020.

\bibitem{Collaco2021}
Elen Colla{\c{c}}o, Elisabeti Kira, Lucas~H. Sallaberry, Anna~C.M. Queiroz, Maria~A.A.M. Machado, Oswaldo Crivello, and Romero Tori.
\newblock {Immersion and haptic feedback impacts on dental anesthesia technical skills virtual reality training}.
\newblock {\em Journal of Dental Education}, 85(4):589--598, apr 2021.

\bibitem{Gani2022}
Abrar Gani, Oliver Pickering, Caroline Ellis, Omar Sabri, and Philip Pucher.
\newblock {Impact of haptic feedback on surgical training outcomes: A Randomised Controlled Trial of haptic versus non-haptic immersive virtual reality training}.
\newblock {\em Annals of Medicine and Surgery}, 83:104734, 2022.

\bibitem{Chrysanthakopoulou2021}
Agapi Chrysanthakopoulou, Konstantinos Kalatzis, and Konstantinos Moustakas.
\newblock {Immersive Virtual Reality Experience of Historical Events Using Haptics and Locomotion Simulation}.
\newblock {\em Applied Sciences 2021, Vol. 11, Page 11613}, 11(24):11613, dec 2021.

\bibitem{Edwards2019}
Bosede~Iyiade Edwards, Kevin~S. Bielawski, Rui Prada, and Adrian~David Cheok.
\newblock {Haptic virtual reality and immersive learning for enhanced organic chemistry instruction}.
\newblock {\em Virtual Reality}, 23(4):363--373, dec 2019.

\bibitem{Bortone2018}
Ilaria Bortone, Daniele Leonardis, Nicola Mastronicola, Alessandra Crecchi, Luca Bonfiglio, Caterina Procopio, Massimiliano Solazzi, and Antonio Frisoli.
\newblock {Wearable Haptics and Immersive Virtual Reality Rehabilitation Training in Children with Neuromotor Impairments}.
\newblock {\em IEEE Transactions on Neural Systems and Rehabilitation Engineering}, 26(7):1469--1478, jul 2018.

\bibitem{chattha2020motion}
Umer~Asghar Chattha, Uzair~Iqbal Janjua, Fozia Anwar, Tahir~Mustafa Madni, Muhammad~Faisal Cheema, and Sana~Iqbal Janjua.
\newblock Motion sickness in virtual reality: An empirical evaluation.
\newblock {\em IEEE Access}, 8:130486--130499, 2020.

\bibitem{kolasinski1995simulator}
Eugenia~M Kolasinski.
\newblock Simulator sickness in virtual environments.
\newblock 1995.

\bibitem{lo2001cybersickness}
WT~Lo and Richard~HY So.
\newblock Cybersickness in the presence of scene rotational movements along different axes.
\newblock {\em Applied ergonomics}, 32(1):1--14, 2001.

\bibitem{matt2021}
Matthias Nürnberger, Carsten Klingner, Otto~W. Witte, and Stefan Brodoehl.
\newblock Mismatch of visual-vestibular information in virtual reality: Is motion sickness part of the brains attempt to reduce the prediction error?
\newblock {\em Frontiers in Human Neuroscience}, 15, 2021.

\bibitem{NG2020101922}
Adrian~K.T. Ng, Leith~K.Y. Chan, and Henry~Y.K. Lau.
\newblock A study of cybersickness and sensory conflict theory using a motion-coupled virtual reality system.
\newblock {\em Displays}, 61:101922, 2020.

\bibitem{Heo2020}
Jaeseok Heo and Gilwon Yoon.
\newblock {EEG Studies on Physical Discomforts Induced by Virtual Reality Gaming}.
\newblock {\em Journal of Electrical Engineering and Technology}, 15(3):1323--1329, may 2020.

\bibitem{Chang2020}
Eunhee Chang, Hyun~Taek Kim, and Byounghyun Yoo.
\newblock {Virtual Reality Sickness: A Review of Causes and Measurements}.
\newblock {\em International Journal of Human–Computer Interaction}, 36(17):1658--1682, oct 2020.

\bibitem{Langbehn2018}
Eike Langbehn, Paul Lubos, and Frank Steinicke.
\newblock {Evaluation of locomotion techniques for room-scale VR: Joystick, teleportation, and redirected walking}.
\newblock {\em ACM International Conference Proceeding Series}, (18), apr 2018.

\bibitem{Choi2020}
InBeom Choi, Jong-Jin Park, ShinWoo Kim, and Hyung-Chul~O Li.
\newblock {Effect of Inconsistency Between Visually Perceived Walking Speed and Physically Perceived Walking Speed on VR Sickness in VR-Treadmill Walking}.
\newblock {\em Science of Emotion and Sensibility}, 23(3):79--90, sep 2020.

\bibitem{Mayor2021}
Jesus Mayor, Laura Raya, and Alberto Sanchez.
\newblock {A Comparative Study of Virtual Reality Methods of Interaction and Locomotion Based on Presence, Cybersickness, and Usability}.
\newblock {\em IEEE Transactions on Emerging Topics in Computing}, 9(3):1542--1553, 2021.

\bibitem{Cherni2021}
Heni Cherni, Souliman Nicolas, and Natacha M{\'{e}}tayer.
\newblock {Using virtual reality treadmill as a locomotion technique in a navigation task: Impact on user experience – case of the KatWalk}.
\newblock {\em International Journal of Virtual Reality}, 21(1):1--14, may 2021.

\bibitem{Kitson2017}
Alexandra Kitson, Abraham~M. Hashemian, Ekaterina~R. Stepanova, Ernst Kruijff, and Bernhard~E. Riecke.
\newblock {Comparing leaning-based motion cueing interfaces for virtual reality locomotion}.
\newblock {\em 2017 IEEE Symposium on 3D User Interfaces, 3DUI 2017 - Proceedings}, pages 73--82, apr 2017.

\bibitem{Bovim2020}
Lars~Peder Bovim, Beate~Eltarvag Gjesdal, Silje Maeland, Mona~K. Aaslund, and Bard Bogen.
\newblock {The impact of motor task and environmental constraints on gait patterns during treadmill walking in a fully immersive virtual environment}.
\newblock {\em Gait \& Posture}, 77:243--249, mar 2020.

\bibitem{Vukelic2023}
Goran Vukelic, Dario Ogrizovic, Dean Bernecic, Darko Glujic, and Goran Vizentin.
\newblock {Application of VR Technology for Maritime Firefighting and Evacuation Training—A Review}.
\newblock {\em Journal of Marine Science and Engineering 2023, Vol. 11, Page 1732}, 11(9):1732, sep 2023.

\bibitem{Conges2020}
Aurelie Conges, Alexis Evain, Frederick Benaben, Olivier Chabiron, and Sebastien Rebiere.
\newblock {Crisis Management Exercises in Virtual Reality}.
\newblock {\em Proceedings - 2020 IEEE Conference on Virtual Reality and 3D User Interfaces, VRW 2020}, pages 87--92, mar 2020.

\bibitem{Boletsis2017}
Costas Boletsis.
\newblock The new era of virtual reality locomotion: A systematic literature review of techniques and a proposed typology.
\newblock {\em Multimodal Technologies and Interaction 2017, Vol. 1, Page 24}, 1:24, 9 2017.

\bibitem{Bozgeyikli2019}
Evren Bozgeyikli, Andrew Raij, Srinivas Katkoori, and Rajiv Dubey.
\newblock Locomotion in virtual reality for room scale tracked areas.
\newblock {\em International Journal of Human-Computer Studies}, 122:38--49, 2 2019.

\bibitem{Cherni2020}
Heni Cherni, Natacha Métayer, and Nicolas Souliman.
\newblock Literature review of locomotion techniques in virtual reality.
\newblock {\em International Journal of Virtual Reality}, 20:1--20, 3 2020.

\bibitem{Razzaque2001}
Sharif Razzaque, Zachariah Kohn, and Mary~C. Whitton.
\newblock {Redirected Walking}.
\newblock In {\em Eurographics 2001 - Short Presentations}. Eurographics Association, 2001.

\bibitem{Matsumoto2016}
Keigo Matsumoto, Yuki Ban, Takuji Narumi, Yohei Yanase, Tomohiro Tanikawa, and Michitaka Hirose.
\newblock Unlimited corridor: Redirected walking techniques using visuo haptic interaction.
\newblock {\em ACM SIGGRAPH 2016 Emerging Technologies, SIGGRAPH 2016}, 7 2016.

\bibitem{Steinicke2010}
Frank Steinicke, Gerd Bruder, Jason Jerald, Harald Frenz, and Markus Lappe.
\newblock Estimation of detection thresholds for redirected walking techniques.
\newblock {\em IEEE Transactions on Visualization and Computer Graphics}, 16:17--27, 1 2010.

\bibitem{Slater1995}
Mel Slater, Anthony Steed, and Martin Usoh.
\newblock The virtual treadmill: A naturalistic metaphor for navigation in immersive virtual environments.
\newblock pages 135--148, 1995.

\bibitem{Iwata2000}
Hiroo Iwata.
\newblock Locomotion interface for virtual environments.
\newblock {\em Robotics Research}, pages 275--282, 2000.

\bibitem{Hager}
Holger Hager, Tuncay Cakmak, Johara J{\"a}gers, and Cyberith GmbH.
\newblock Cyberith virtualizer elite 2 – second generation vr locomotion device based on a 2 dof motion platform.
\newblock 2019.

\bibitem{barkan201400}
Andrew Barkan, Jeffrey Skidmore, and Panagiotis Artemiadis.
\newblock {Variable Stiffness Treadmill (VST): A novel tool for the investigation of gait}.
\newblock {\em Proceedings - IEEE International Conference on Robotics and Automation}, pages 2838--2843, sep 2014.

\bibitem{Skidmore2015}
Jeffrey Skidmore, Andrew Barkan, and Panagiotis Artemiadis.
\newblock {Variable Stiffness Treadmill (VST): System Development, Characterization, and Preliminary Experiments}.
\newblock {\em IEEE/ASME Transactions on Mechatronics}, 20(4):1717--1724, aug 2015.

\bibitem{hernandez201800}
Ernesto Hernandez, Christian Warhmund, Kyle Lamoureux, Essie Lee, Isaac Sanchez, Whitney Matthews, and Amir Jafari.
\newblock A novel treadmill that can bilaterally adjust the vertical surface stiffness.
\newblock {\em IEEE/ASME Transactions on Mechatronics}, 23(5):2338--2346, 2018.

\bibitem{Pyo2018}
Sang~Hun Pyo, Ho~Su Lee, Bui~Minh Phu, Sang~Joon Park, and Jung~Won Yoon.
\newblock Development of an fast-omnidirectional treadmill (f-odt) for immersive locomotion interface.
\newblock {\em 2018 IEEE International Conference on Robotics and Automation (ICRA)}, pages 760--766, 2018.

\bibitem{Hollerbach2000}
John~M. Hollerbach, Yangming Xu, Robert~R. Christensen, and Stephen~C. Jacobsen.
\newblock {Design Specifications for the Second Generation Sarcos Treadport Locomotion Interface}.
\newblock {\em ASME International Mechanical Engineering Congress and Exposition, Proceedings (IMECE)}, 2000-O(2):1293--1298, nov 2021.

\bibitem{schmidt2005hapticwalker}
Henning Schmidt, Stefan Hesse, Rolf Bernhardt, and J\"{o}rg Kr\"{u}ger.
\newblock Hapticwalker---a novel haptic foot device.
\newblock {\em ACM Trans. Appl. Percept.}, 2(2):166–180, apr 2005.

\bibitem{boian2005}
R.F. Boian, M.~Bouzit, G.C. Burdea, J.~Lewis, and J.E. Deutsch.
\newblock Dual stewart platform mobility simulator.
\newblock In {\em 9th International Conference on Rehabilitation Robotics, 2005. ICORR 2005.}, pages 550--555, 2005.

\bibitem{Iwata20010}
H.~Iwata, H.~Yano, and F.~Nakaizumi.
\newblock {Gait Master: A versatile locomotion interface for uneven virtual terrain}.
\newblock {\em Proceedings - Virtual Reality Annual International Symposium}, pages 131--137, 2001.

\bibitem{yoon2010planar}
Jungwon Yoon, Jangwoo Park, and Jeha Ryu.
\newblock A planar symmetric walking cancellation algorithm for a foot—platform locomotion interface.
\newblock {\em The International Journal of Robotics Research}, 29(1):39--59, 2010.

\bibitem{yoon2010}
Jungwon Yoon, Bondhan Novandy, Chul-Ho Yoon, and Ki-Jong Park.
\newblock A 6-dof gait rehabilitation robot with upper and lower limb connections that allows walking velocity updates on various terrains.
\newblock {\em IEEE/ASME Transactions on Mechatronics}, 15(2):201--215, 2010.

\bibitem{laplante2004continuing}
John~N Laplante and Thomas~P Kaeser.
\newblock The continuing evolution of pedestrian walking speed assumptions.
\newblock {\em Institute of Transportation Engineers. ITE Journal}, 74(9):32, 2004.

\bibitem{keemink2018admittance}
Arvid~QL Keemink, Herman van~der Kooij, and Arno~HA Stienen.
\newblock Admittance control for physical human--robot interaction.
\newblock {\em The International Journal of Robotics Research}, 37(11):1421--1444, 2018.

\bibitem{colgate1994}
J.~Edward Colgate and Gerd Schenkel.
\newblock {Passivity of a class of sampled-data systems: application to haptic interfaces}.
\newblock {\em Proceedings of the American Control Conference}, 3:3236--3240, 1994.

\bibitem{o2009improved}
Marcia~K O’Malley, Kevin~S Sevcik, and Emilie Kopp.
\newblock Improved haptic fidelity via reduced sampling period with an fpga-based real-time hardware platform.
\newblock {\em Journal of computing and information science in engineering}, 9(1), 2009.

\bibitem{ben2022}
Benjamin Beiter, Connor Herron, and Alexander Leonessa.
\newblock Whole body control for haptic interaction with vr.
\newblock In {\em 2022 American Control Conference (ACC)}, pages 653--658, 2022.

\bibitem{kurfess2005robotics}
Thomas~R Kurfess et~al.
\newblock {\em Robotics and automation handbook}, volume 414.
\newblock CRC press Boca Raton, FL, 2005.

\bibitem{yoshikawa1995}
Tsuneo Yoshikawa, Yasuyoshi Yokokohji, Tomoharu Matsumoto, and Xin-Zhi Zheng.
\newblock {Display of Feel for the Manipulation of Dynamic Virtual Objects}.
\newblock {\em Journal of Dynamic Systems, Measurement, and Control}, 117(4):554--558, 12 1995.

\bibitem{Lecours2012}
Alexandre Lecours, Boris Mayer-St-Onge, and Cl{\'{e}}ment Gosselin.
\newblock {Variable admittance control of a four-degree-of-freedom intelligent assist device}.
\newblock {\em Proceedings - IEEE International Conference on Robotics and Automation}, (2):3903--3908, 2012.

\bibitem{Yoon2010a}
Jungwon Yoon, Bondhan Novandy, Chul~Ho Yoon, and Ki~Jong Park.
\newblock {A 6-DOF gait rehabilitation robot with upper and lower limb connections that allows walking velocity updates on various terrains}.
\newblock {\em IEEE/ASME Transactions on Mechatronics}, 15(2):201--215, apr 2010.

\bibitem{Yoon2009}
Jungwon Yoon, Jangwoo Park, and Jeha Ryu.
\newblock {A Planar Symmetric Walking Cancellation Algorithm for a Foot—Platform Locomotion Interface}.
\newblock {\em http://dx.doi.org/10.1177/0278364909104293}, 29(1):39--59, may 2009.

\bibitem{Kim2023}
Jongbum Kim, Seunghue Oh, Yongjin Jo, James~Hyungsup Moon, and Jonghyun Kim.
\newblock {A robotic treadmill system to mimic overground walking training with body weight support}.
\newblock {\em Frontiers in Neurorobotics}, 17, jun 2023.

\bibitem{Dong2010}
Haiwei Dong, Tatuo Oshiumi, Akinori Nagano, and Zhiwei Luo.
\newblock {Development of a 3D interactive virtual market system with adaptive treadmill control}.
\newblock {\em IEEE/RSJ 2010 International Conference on Intelligent Robots and Systems, IROS 2010 - Conference Proceedings}, pages 5238--5244, 2010.

\bibitem{Yu2021}
{Principal Component Analysis of the Running Ground Reaction Forces With Different Speeds}.
\newblock {\em Frontiers in Bioengineering and Biotechnology}, 9:629809, mar 2021.

\bibitem{Feasel2011}
Jeff Feasel, Mary~C. Whitton, Laura Kassler, Frederick~P. Brooks, and Michael~D. Lewek.
\newblock The integrated virtual environment rehabilitation treadmill system.
\newblock {\em IEEE Transactions on Neural Systems and Rehabilitation Engineering}, 19:290--297, 6 2011.

\end{thebibliography}

\end{document}